\documentclass{article} 
\usepackage{iclr2026_conference,times}
\usepackage{times}
\usepackage{latexsym}
\usepackage{hyperref}

\usepackage[T1]{fontenc}

\usepackage[utf8]{inputenc}
\usepackage{newunicodechar}
\newunicodechar{−}{\textminus}

\usepackage{microtype}

\usepackage{inconsolata}

\usepackage{graphicx}
\usepackage{tcolorbox}           
\tcbuselibrary{breakable}
\usepackage{enumitem}            
\usepackage{arydshln}
\usepackage{float}
\usepackage{subfigure}
\usepackage{booktabs} 
\usepackage{listings}
\usepackage{xcolor}
\usepackage{textcomp}
\usepackage[ruled,vlined]{algorithm2e}
\usepackage[utf8]{inputenc}
\usepackage{listings}
\lstdefinelanguage{lean}{}
\usepackage{multirow}
\usepackage{mdframed}
\usepackage{algpseudocode}
\usepackage{bbm}
\usepackage[utf8]{inputenc}
\usepackage{epstopdf}
\usepackage{amsthm}  
\usepackage{amsmath} 
\usepackage{amssymb} 
\usepackage[utf8]{inputenc}

\usepackage{adjustbox}
\usepackage{multirow}
\usepackage{colortbl}  
\usepackage{array}
\definecolor{darkgreen}{RGB}{60,151,77}
\usepackage{amsmath,amssymb}
\tcbuselibrary{breakable}

\allowdisplaybreaks

\usepackage{marvosym}  
\usepackage{textcomp}
\usepackage{caption}
\lablogoone{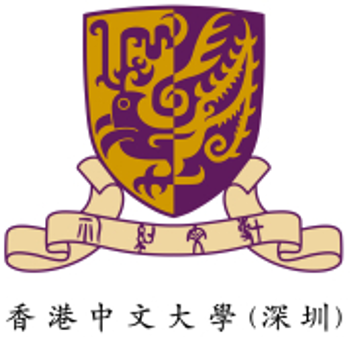} 
\lablogotwo{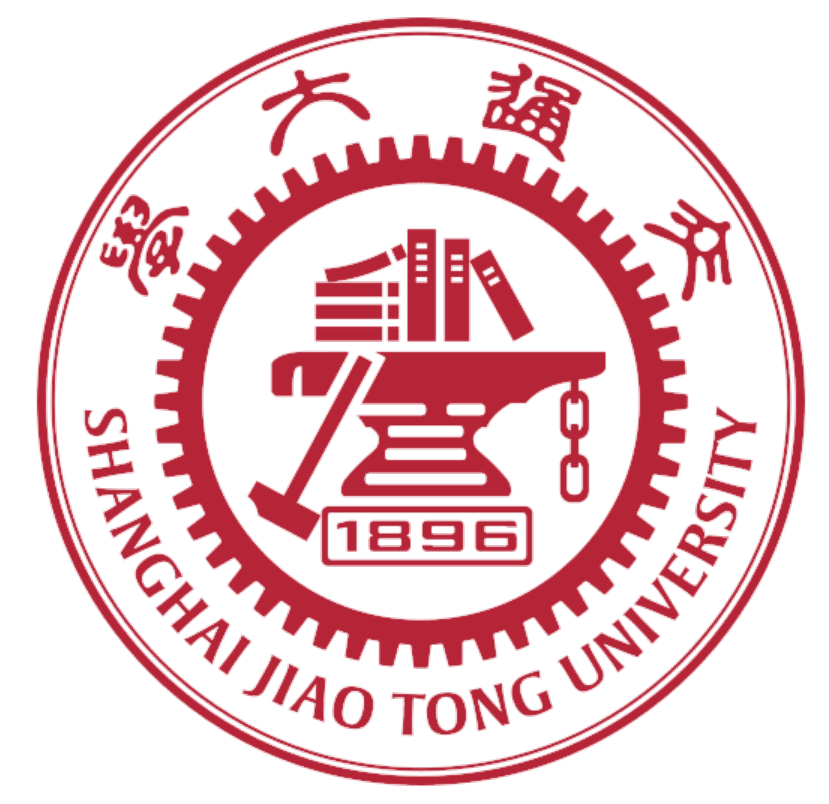} 
\lablogothree{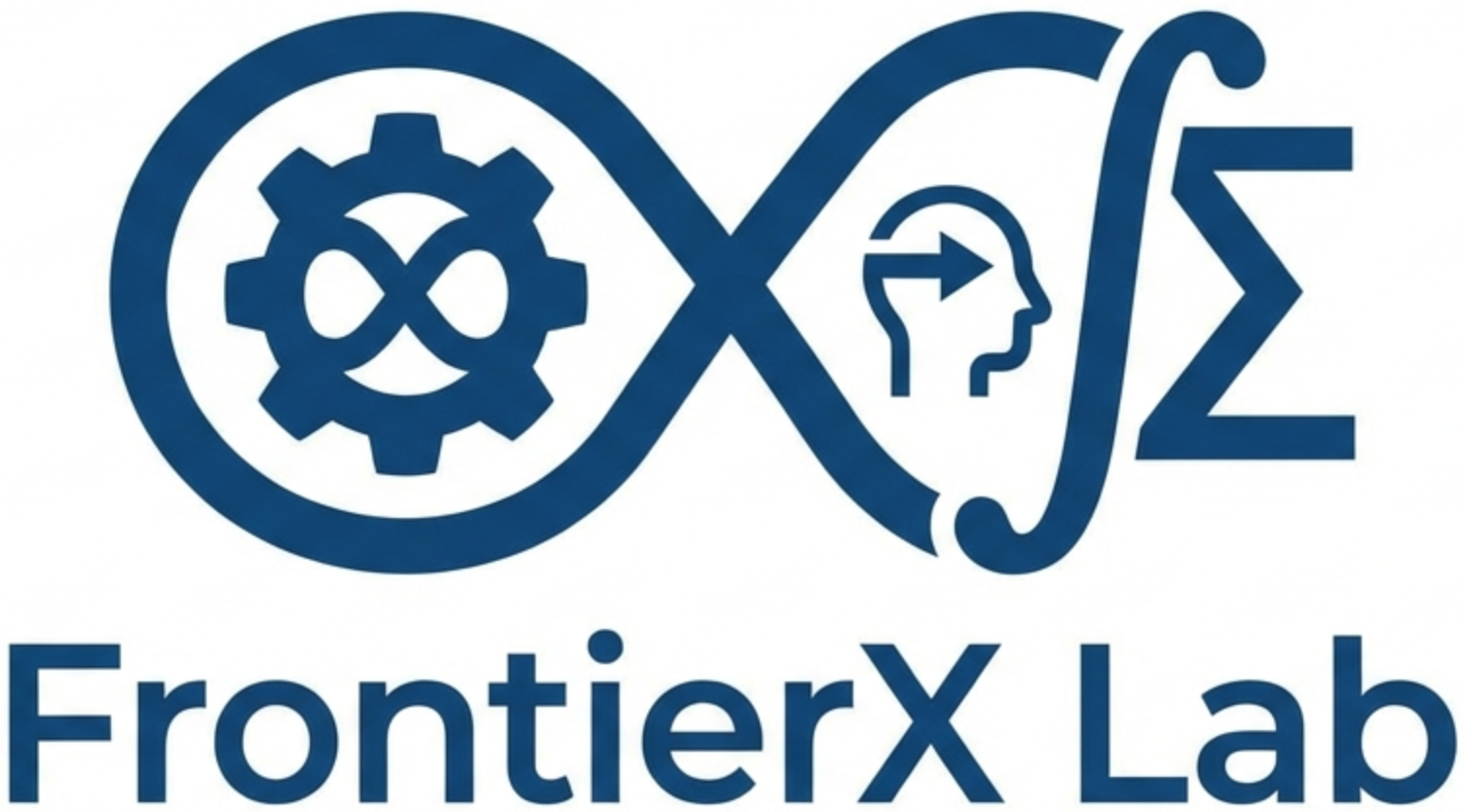} 
\labname{CUHKSZ \& SJTU}  

\title{Milestones over Outcome: Unlocking Geometric Reasoning with Sub-Goal Verifiable Reward}

\author{
 \textbf{Jianlong Chen\textsuperscript{1}},
 \textbf{Daocheng Fu\textsuperscript{3}},
 \textbf{Shengze Xu\textsuperscript{4}},
 \textbf{Jiawei Chen\textsuperscript{5}},
 \textbf{Yuan Feng\textsuperscript{2}},\\
 \textbf{Yue Yang\textsuperscript{2}},
 \textbf{Junchi Yan\textsuperscript{2}},
 \textbf{Hongyuan Zha\textsuperscript{1}},
 \textbf{Renqiu Xia\textsuperscript{2, \Letter}}\\
 \vspace{1em}
 \textmd{\textsuperscript{1}The Chinese University of Hong Kong, Shenzhen
 \textsuperscript{2}Shanghai Jiao Tong University\\
 \textsuperscript{3}Fudan University
 \textsuperscript{4}The Chinese University of Hong Kong \\
 \textsuperscript{5}University of Science and Technology Beijing}\\
\vspace{0.5em}
\normalsize \textmd{\{jianlongchen\}@link.cuhk.edu.cn, \{xiarenqiu\}@sjtu.edu.cn\ \ \ \ \ \textsuperscript{\Letter} \ Corresponding Authors\\
}
}

\begin{document}

\maketitle

\begin{abstract}
\normalfont
Multimodal Large Language Models (MLLMs) struggle with complex geometric reasoning, largely because "black box" outcome-based supervision fails to distinguish between lucky guesses and rigorous deduction. To address this, we introduce a paradigm shift towards subgoal-level evaluation and learning. We first construct \textit{GeoGoal}, a benchmark synthesized via a rigorous formal verification data engine, which converts abstract proofs into verifiable numeric subgoals. This structure reveals a critical divergence between reasoning quality and outcome accuracy. Leveraging this, we propose the \textbf{S}ub-\textbf{G}oal \textbf{V}erifiable \textbf{R}eward (SGVR) framework, which replaces sparse signals with dense rewards based on the \textit{Skeleton Rate}. Experiments demonstrate that SGVR not only enhances geometric performance (\textbf{+9.7\%}) but also exhibits strong generalization, transferring gains to general math (\textbf{+8.0\%}) and other general reasoning tasks (\textbf{+2.8\%}), demonstrating broad applicability across diverse domains.\\

\normalfont
\vspace{0.6em}
\raisebox{-0.2\height}{\includegraphics[height=0.15in]{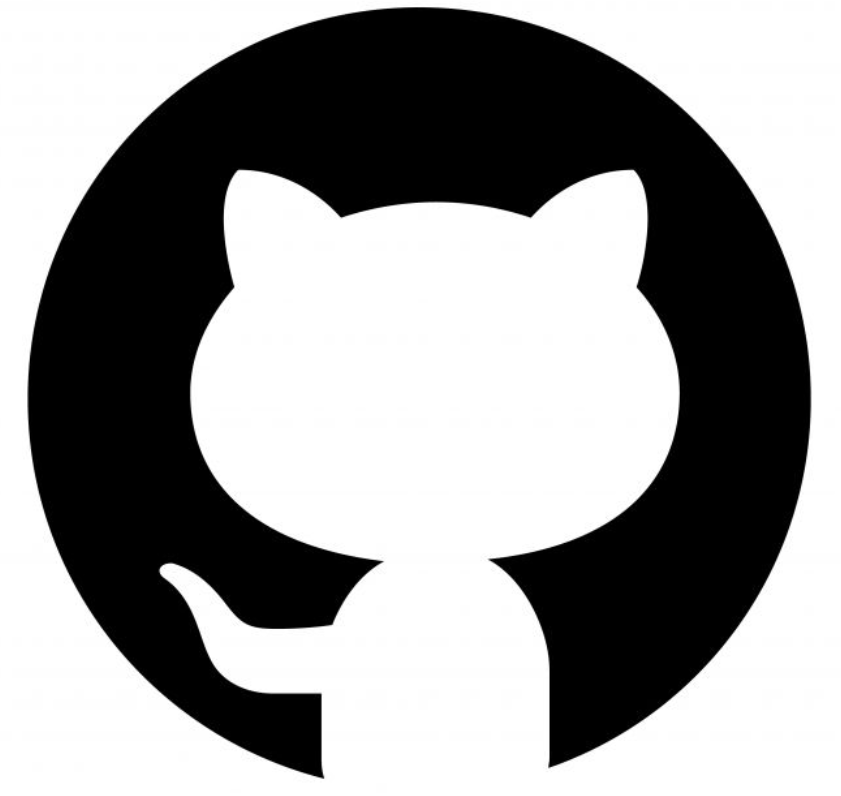}} code: \url{https://github.com/FrontierX-Lab/SGVR}.\\

\end{abstract}

\section{Introduction}

Multimodal Large Language Models (MLLMs) have demonstrated impressive proficiency in diverse vision-language tasks~\cite{achiam2023gpt, team2023gemini,bai2025qwen2}. 
However, their efficacy diminishes in domains requiring rigorous multi-step reasoning. 
Geometric reasoning stands as a formidable frontier, necessitating the coherent integration of visual perception, symbolic abstraction, and logical deduction~\cite{trinh2024solving, he2024olympiadbench}. 
While specialized neuro-symbolic solvers~\cite{trinh2024solving, sicca2024newclid} have reached Olympiad-level performance, general-purpose MLLMs continue to struggle with long-horizon inference, often plagued by hallucinations and logical gaps in natural language. 

\begin{figure}[t]
    \centering
    \includegraphics[width=0.62\textwidth]{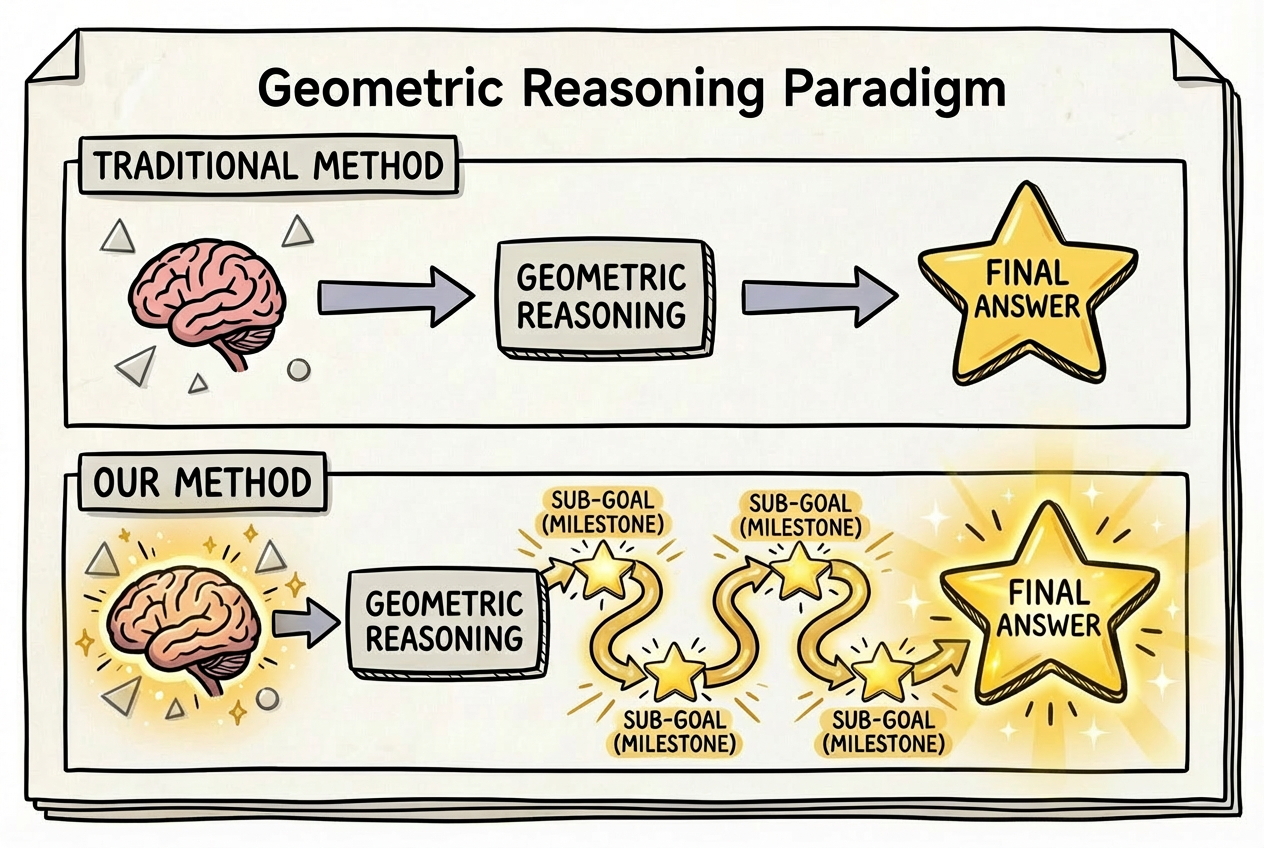}
    \captionof{figure}{\textbf{Our main goal:} Decomposing the "black box" of complex geometric reasoning into a verifiable chain of fine-grained intermediate milestones.
    }
    \vspace{-1em}
    \label{fig:general}
\end{figure}

Standard evaluation benchmarks~\cite{geoqa_ngs/chen2021geoqa, lu2021inter} treat reasoning as a black box, assessing only the final numerical result. 
This coarse objective creates a misalignment between metric and capability: it yields \textit{false positives} via spurious correlations and \textit{false negatives} via minor arithmetic slips. 
Crucially, sparse final-answer signals fail to provide the fine-grained feedback necessary for models to learn robust intermediate deductive steps. 
More fundamentally, outcome accuracy is not a faithful proxy for step-wise reasoning reliability: models can sometimes recover the correct final answer despite flawed intermediate steps, while otherwise valid reasoning can be penalized by small downstream errors.
Our solution is to break open this black box by focusing on the reasoning milestones. As illustrated in Figure~\ref{fig:general}, we reframe the entire process as a sequence of verifiable sub-goals. This structure offers a unified solution for both evaluation and learning: it allows for a granular evaluation of the reasoning path, pinpointing exactly where logic fails, while simultaneously providing the dense, trustworthy signals required for effective training.

\begin{figure*}[t]
    \centering
    \includegraphics[width=0.9\textwidth]{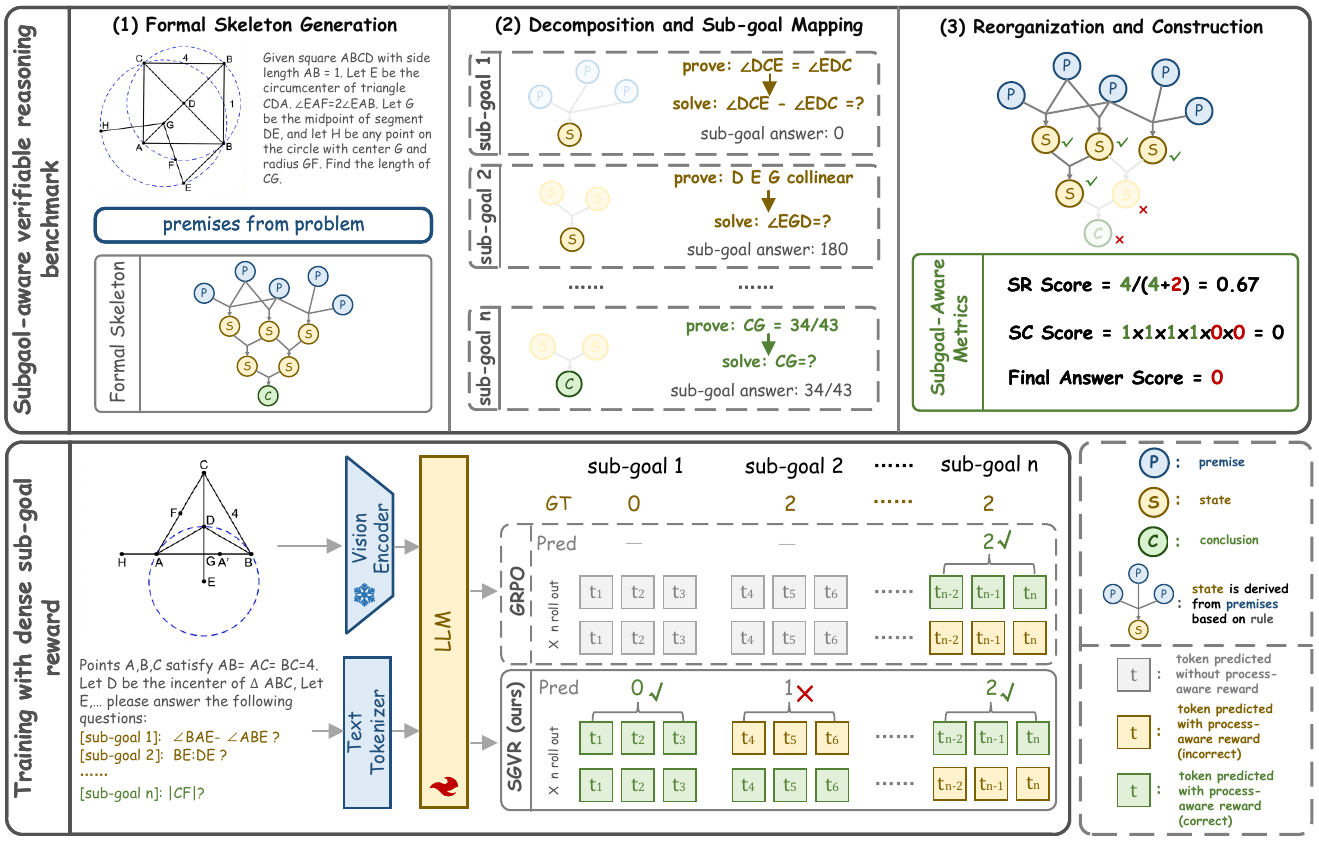}
    \captionof{figure}{\textbf{Overall framework:} 
    (\textbf{Top}) \textit{Benchmark construction}: Formally verified skeletons from TrustGeoGen~\cite{trustgeogen2025} are decomposed into numeric sub-goals to enable subgoal-level metrics. 
    (\textbf{Bottom}) \textit{SGVR training}: The model generates structured traces; predicted sub-goals are verified against ground truth to formulate dense rewards for policy optimization via GRPO.
    }
    \vspace{-1em}
    \label{fig:pipeline}
\end{figure*}

In this work, we introduce a paradigm shift towards \textit{subgoal-level evaluation and reinforcement learning}, as depicted in Figure~\ref{fig:pipeline}. 
First, we create the \textit{GeoGoal} benchmark, synthesized via the TrustGeoGen data engine~\cite{trustgeogen2025}. Our "proofing-to-solving" transformation convert abstract logical predicates into a sequence of executable, verifiable numeric sub-goals. This structures the reasoning process into a series of clear milestones, moving beyond unstructured text generation. Critically, our evaluation using \textit{GeoGoal} reveals that reasoning quality and outcome accuracy can diverge, which motivates the need for more granular, subgoal-level supervision.
To address this gap, we then leverage \textit{GeoGoal} to propose the \textit{Sub-Goal Verifiable Reward (SGVR)} framework. This method facilitates Reinforcement Learning with Verifiable Rewards (RLVR) by replacing sparse outcome rewards with dense, subgoal-oriented signals. Specifically, we use Group Relative Policy Optimization (GRPO)~\cite{deepseek2025deepseek} to maximize the \textit{Skeleton Rate}, the ratio of successfully verified sub-goals. The results show that our proposed SGVR improves both final-answer performance and intermediate reasoning quality, with gains transferring beyond geometry by achieving average improvements of \textcolor{darkgreen}{\textbf{+9.7\%}} on geometric reasoning, \textcolor{darkgreen}{\textbf{+8.0\%}} on general mathematics, and \textcolor{darkgreen}{\textbf{+2.8\%}} on general reasoning tasks.

Our contributions are summarized as follows:
\begin{enumerate}[itemsep=0pt, topsep=0pt, parsep=0pt, partopsep=0pt]
    \item \textbf{Verifiable Benchmark Construction:} We present the first multimodal geometry benchmark \textit{GeoGoal} where intermediate sub-goals are formally verified and automatically checkable, introducing Skeleton Rate (SR), Skeleton Completion (SC) and Consistency Ratio (CR) as rigorous metrics for reasoning fidelity.
    \item \textbf{SGVR Framework:} We propose a reinforcement learning framework leveraging verifiable numeric sub-goals as critical reasoning milestones to provide dense supervision.
    \item \textbf{Empirical Efficacy:} Experiments show that our proposed SGVR framework improves final answer accuracy with robust cross-domain transfer to general reasoning tasks and enhances intermediate reasoning quality.
\end{enumerate}

\section{GeoGoal: A Verifiable Benchmark for Sub-Goal Reasoning}
\label{sec:benchmark}

Standard geometry benchmarks, which evaluate only final answers, cannot distinguish between genuine reasoning and heuristic shortcuts. This paradigm also precludes subgoal-level evaluation signals, which are essential for training models that reason robustly. To overcome these limitations, we construct a benchmark designed for \textit{milestones verifiability}, where each reasoning step has a verifiable ground truth, and \textit{fine-grained subgoal assessment}.

\subsection{Construction Pipeline}
\label{sec:construction}

Our pipeline, illustrated in Figure~\ref{fig:pipeline}, transforms formal proofs into a benchmark with verifiable sub-goals, enabling dense reward signals for downstream RL. The overall procedure is organized as:

\paragraph{Step 1: Data Engine: Formal Skeleton Generation}
We leverage TrustGeoGen~\citep{trustgeogen2025} to synthesize complete formal problem instances. For each sample, the engine outputs the geometric premise together with a formally verified solution skeleton $\{\mathcal{S}_t\}_{t=1}^{T}$. A built-in verifier enforces type constraints, dependency ordering, and derivability of every predicate, ensuring the logical correctness of the reasoning chain.

\paragraph{Step 2: Decomposition and Sub-goal Mapping}
To enable step-wise verification, we first decompose the formal solution skeleton $\{\mathcal{S}_t\}$ into atomic reasoning steps. Since formal predicates are abstract and not directly solvable by standard LLMs, we map each decomposed predicate into a numeric sub-goal via a mapping function\footnote{The complete mapping rules can be found in Tables~\ref{tab:mapping_core}, ~\ref{tab:mapping_constants}, ~\ref{tab:mapping_triangles}, ~\ref{tab:mapping_quads} and~\ref{tab:mapping_constructions}}. For example, a congruence predicate \texttt{cong(A,B,C,D)} is mapped to a length ratio task $(\mathcal{T}_t: |AB|/|CD|,\; y_t: 1)$. 
This conversion turns abstract reasoning steps into automatically checkable numeric targets.

\paragraph{Step 3: Reorganization and Sequence Construction}
Finally, we reorganize these sub-goals into a sequential format to construct the final benchmark instances. 
Crucially, the sequence is ordered such that the \textit{last sub-goal} $\mathcal{T}_{n_i}$ corresponds to the original problem's final goal, while preceding sub-goals represent intermediate reasoning steps derived from the formal proof.
The model is presented with the initial problem and premises, and is required to find the values for the entire sequence of sub-goals. 
This structure allows us to verify the model's reasoning process step-by-step, rather than just checking the final answer.

\subsection{Sub-goal Evaluation Metrics}
\label{sec:eval_metrics}

To capture performance at different granularities, we consider three complementary metrics.
Skeleton Rate (SR) measures average step-wise correctness across sub-goals,
Skeleton Completion (SC) measures end-to-end consistency over complete reasoning chains,
and the Consistency Ratio (CR) quantifies the \emph{normalized} alignment between the two at the dataset level,
i.e., how much subgoal-wise correctness translates into fully consistent solutions.
CR is computed as the ratio of the dataset-level SC to the dataset-level SR.
For instance $i$ with $n_i$ sub-goals, let
$p_i=\frac{1}{n_i}\sum_{t=1}^{n_i}\mathbb{I}(\hat{y}_{i,t}=y_{i,t})$
denote the fraction of correctly solved sub-goals, and
$c_i=\prod_{t=1}^{n_i}\mathbb{I}(\hat{y}_{i,t}=y_{i,t})$
indicate whether \emph{all} sub-goals are correct.
By construction, $c_i \le p_i$ for all $i$, hence $\text{SC} \le \text{SR}$.
\vspace{-5pt}
\begin{equation}
\small
\begin{aligned}
\text{SR}
&= \frac{1}{N}\sum_{i=1}^{N} \frac{1}{n_i}\sum_{t=1}^{n_i}
\mathbb{I}(\hat{y}_{i,t} = y_{i,t}), \\
\text{SC}
&= \frac{1}{N}\sum_{i=1}^{N}
\prod_{t=1}^{n_i}\mathbb{I}(\hat{y}_{i,t} = y_{i,t}), \\
\text{CR}
&=
\begin{cases}
\frac{\text{SC}_{\text{dataset}}}{\text{SR}_{\text{dataset}}}, & \text{if } \text{SR}_{\text{dataset}} > 0,\\
0, & \text{otherwise.}
\end{cases}
\end{aligned}
\end{equation}

Intuitively, $\text{CR}$ can be viewed as an IoU-like consistency ratio,
measuring the fraction of step-wise correctness that also forms fully correct chains;
larger values indicating stronger reasoning stability and less error propagation along reasoning chains.

\subsection{Dataset Characteristics}

We construct balanced Train and Test splits of 256 instances each. The test set is intentionally skewed toward longer reasoning chains to probe generalization beyond the training distribution. Each instance contributes multiple verifiable sub-goals, yielding dense signals for both evaluation and RL training. For detailed proof-length distributions and geometric concept coverage, please refer to Appendix~\ref{app:dataset_stats}.

\subsection{Benchmark Evaluation of Existing Models}
\label{sec:benchmark_eval}

Beyond serving as training data, our benchmark also enables subgoal-level evaluation of existing multimodal models.
We evaluate some widely deployed models spanning both proprietary and open-weight systems using SR and SC, and also report the standard outcome metric of Final Answer accuracy (FA), i.e., correctness of the last sub-goal (the original final goal).
This analysis provides subgoal baselines and directly tests the central premise highlighted in the introduction: \emph{final-answer accuracy alone is not a faithful proxy for the rigor and integrity of intermediate deductions.}.

\paragraph{Sub-goal baselines.}
Table~\ref{tab:bench_perf} reports SR, SC, CR, and FA accuracy, establishing reference points for step-wise correctness (SR), end-to-end consistency (SC), consistency ratio (CR), and outcome accuracy (FA).
Across models, SC is consistently lower than SR and typically lower than FA, reflecting the strictness of requiring \emph{all} intermediate sub-goals to be correct.


\begin{table}[t]
\centering
\small
\caption{Performance of ten multimodal models on our benchmark.
All metrics are reported in \%.
Gemini~2.5~Pro leads in performance, while skeleton-based metrics reveal differences between per-step correctness (SR), end-to-end consistency (SC), consistency ratio (CR), and outcome-based accuracy (FA). The best and second-best performances were highlighted using \textbf{bold} and \underline{underline}, respectively.}
\setlength{\tabcolsep}{4pt}
\begin{adjustbox}{width=0.68\linewidth}
\begin{tabular}{lcccc}
\hline
Model & SR $\uparrow$ & SC $\uparrow$ & CR $\uparrow$ & FA $\uparrow$ \\
\hline
\multicolumn{5}{l}{\textbf{Closed-source models}} \\
\hline
Gemini 2.5 Pro~\cite{comanici2025gemini}              & \textbf{88.7} & \textbf{44.5} & \textbf{50.2} & 82.8 \\
o4-mini~\cite{openai2025o4mini}                      & \underline{88.3} & \underline{37.1} & \underline{42.0} & \textbf{84.0} \\
GPT-5-mini~\cite{openai2025gpt5systemcard}           & 86.4 & 29.7 & 34.4 & \underline{78.5} \\
GPT-4o~\cite{openai2024gpt4osystemcard}              & 47.9 & 4.7 & 9.8 & 36.3 \\
\hline
\multicolumn{5}{l}{\textbf{Open-source models}} \\
\hline
Qwen2.5-VL-32B-Instruct~\cite{bai2025qwen2} & \textbf{63.3} & 3.9 & 6.2 & \textbf{52.3} \\
Qwen2.5-VL-7B-Instruct~\cite{bai2025qwen2}  & \underline{50.5} & 2.3 & 4.6 & \underline{43.4} \\
Qwen3-VL-8B-Instruct~\cite{qwen2025qwen3vl}          & 48.3 & 17.2 & 35.6 & 37.5 \\
Qwen3-VL-30B-A3B-Instruct~\cite{qwen2025qwen3vl}     & 46.9 & \underline{18.8} & \underline{40.1} & 36.7 \\
Qwen3-VL-8B-Thinking~\cite{qwen2025qwen3vl}          & 44.8 & \textbf{31.6} & \textbf{70.5} & 42.6 \\
Llama-4-Scout~\cite{meta2025llama4scout}             & 12.4 & 0.8 & 6.5 & 10.6 \\
\hline
\end{tabular}
\label{tab:bench_perf}
\end{adjustbox}
\end{table}

\begin{figure}[t]
\centering
\includegraphics[width=0.68\columnwidth]{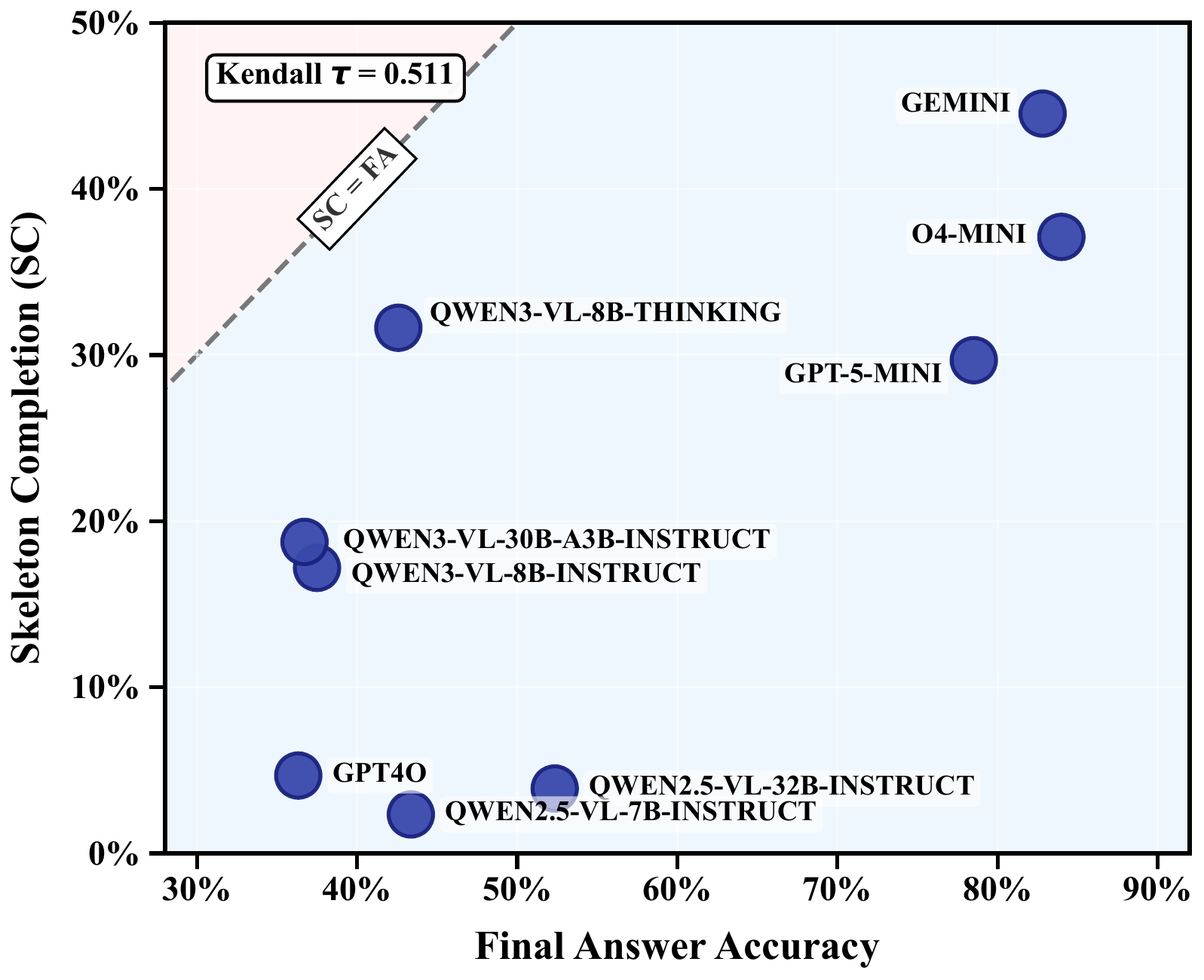}
\caption{Skeleton Completion (SC) versus Final Answer accuracy on our benchmark.
Each point denotes a multimodal model. The light blue background indicates SC < FA. The closer the model is to the line where SC = FA, the more rigorous its reasoning logic is.
}
\label{fig:final_answer_vs_sc}
\end{figure}

\paragraph{How aligned are outcome accuracy and step-wise consistency?}
We compare SC against Final Answer accuracy for all models (Figure~\ref{fig:final_answer_vs_sc}).
The relationship is only moderately aligned (Kendall $\tau = 0.511$): multiple models achieve relatively high Final Answer accuracy despite substantially lower SC.
This divergence implies that outcome-only evaluation can overestimate reasoning reliability, since correct final answers may be produced even when intermediate sub-goals contain errors.

\begin{figure}[t]
\centering
\includegraphics[width=0.68\columnwidth]{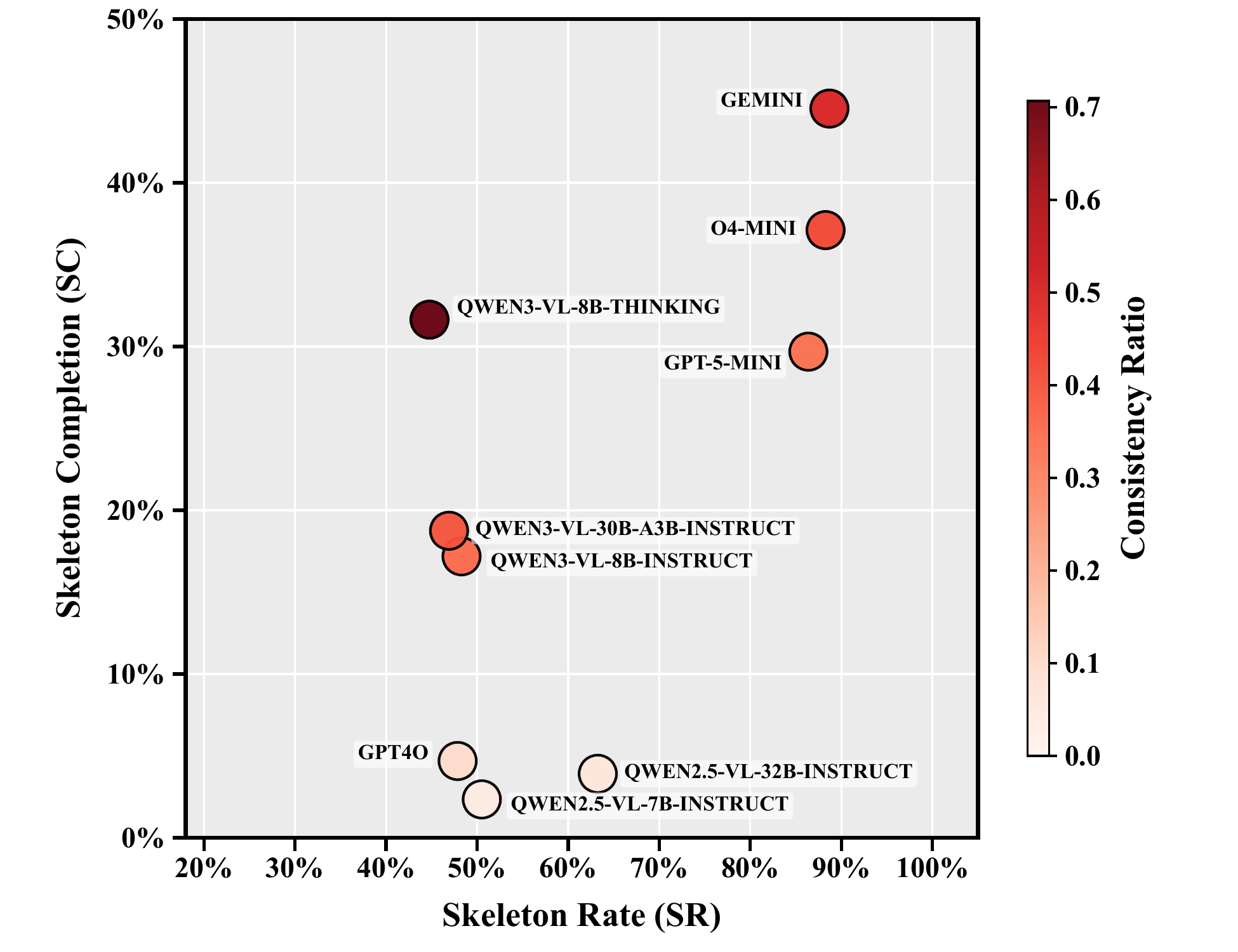}
\caption{Skeleton Completion (SC) v.s. Skeleton Rate (SR) on our benchmark.
Points are color-coded by the Consistency Ratio (CR), revealing distinct trade-offs between step-wise correctness and end-to-end consistency.}
\label{fig:sr_sc_gap_analysis}
\end{figure}

\paragraph{What failure modes are exposed by the relationship between SR and SC?}
We further analyze the joint distribution of SR and SC by plotting models in a two-dimensional space with SR and SC as axes, color-coding each point by its Consistency Ratio (CR) (Figure~\ref{fig:sr_sc_gap_analysis}).
While stronger models tend to cluster toward high SR and high SC, CR still varies substantially, revealing different failure modes.
In particular, models with high SR but low SC exhibit low CR: they solve many individual sub-goals correctly yet fail to maintain end-to-end consistency, suggesting error propagation along long reasoning chains.
By contrast, models with larger CR are more stable, as step-wise correctness is more consistently reflected in complete-chain success.
Together, SR and SC offer complementary diagnostic signals that cannot be inferred from FA alone, motivating their use in both evaluation and dense reward training.

\section{Sub-Goal Verifiable Reward}
\label{sec:method}

Given the step-wise verifiable benchmark in Section~\ref{sec:benchmark}, we introduce \textit{Sub-Goal Verifiable Reward (SGVR)}, a training strategy that exploits automatically checkable sub-goals to produce dense feedback.
As illustrated in the training part of Figure~\ref{fig:pipeline}, the model generates a structured response in which each slot corresponds to a specific sub-goal; every predicted sub-goal is then verified against ground truth, and the resulting verification signals are aggregated into rewards.

\subsection{RL Formulation}

We frame multi-step reasoning as a contextual bandit problem where the generation process is decomposed into a sequence of verifiable sub-goals.
Given a problem $x$, the policy $\pi_\theta$ generates a structured response $y$, which we partition into segments corresponding to individual sub-goals.

\paragraph{Sub-goal Reward Signal.}
A key innovation of SGVR is the construction of a dense reward signal from verifiable intermediate reasoning steps. Unlike outcome-based rewards that only evaluate the final answer, or learned reward models that may hallucinate, our reward is derived from the strict verification of each sub-goal in the reasoning chain. 

For each sub-goal $t$ in a reasoning trajectory with $n$ sub-goals, we define an \textit{intermediate verification}:
\begin{equation}
\small
r_t = \mathbb{I}(\text{verify}(\hat{y}_t, y_t))
\end{equation}
where $\hat{y}_t$ is the predicted value for the $t$-th sub-goal, $y_t$ is the ground truth, and $\mathbb{I}(\text{verify}(\hat{y}_t, y_t))$ indicates whether the prediction matches the verifiable ground truth.

The reward for a complete trajectory is computed as the normalized accumulation of these intermediate signals, which is mathematically equivalent to the instance-level \textit{Skeleton Rate (SR)} metric defined in Section~\ref{sec:eval_metrics}:
\begin{equation}
\small
\mathcal{R}(y) = \text{SR}_i = \frac{1}{n_i} \sum_{t=1}^{n_i} \mathbb{I}(\hat{y}_{i,t} = y_{i,t})
\end{equation}
This formulation is intrinsically \textit{subgoal-level}: rather than a single binary outcome, the reward emerges from the \textit{accumulation} of verification signals throughout the reasoning chain. A trajectory that correctly solves 80\% of the sub-goals receives a significantly higher reward than one that solves only 20\%, even if both fail the final answer. This dense, gradient-like signal provides step-by-step supervision that guides the model to incrementally improve its reasoning process.

\subsection{Group Relative Policy Optimization}

To efficiently optimize the policy using this subgoal-level reward, we employ \textbf{Group Relative Policy Optimization (GRPO)}~\citep{deepseek2025deepseek}. GRPO eliminates the need for a separate value function critic, which is often computationally expensive and unstable to train, by leveraging the group-based relative advantage.

For each question $q$, we sample a group of $G$ outputs $\{o_1, o_2, \dots, o_G\}$ from the old policy $\pi_{\theta_{old}}$. For each output $o_i$, we compute the reward $r_i$ using our verifiable subgoal reward function. The advantage $A_i$ for each output is then computed by normalizing the rewards within the group:
\begin{equation}
\small
A_i = \frac{r_i - \text{mean}(\{r_1, \dots, r_G\})}{\text{std}(\{r_1, \dots, r_G\}) + \epsilon}
\end{equation}
where $\epsilon$ is a small constant for numerical stability.

The GRPO objective function is defined as:

\vspace{-2mm}
\begin{equation}\label{eq:grpo_def}
\small
\left\{
\begin{aligned}
\rho_i
&= \frac{\pi_\theta(o_i \mid q)}{\pi_{\theta_{\text{old}}}(o_i \mid q)} \\
\hat{L}_i(\theta)
&= \min\!\Big(
\rho_i A_i,\,
\operatorname{clip}(\rho_i, 1-\epsilon, 1+\epsilon)\, A_i
\Big) \\
\mathcal{L}(\theta)
&= \mathbb{E}_{q,\,o}\!\left[
\frac{1}{G} \sum_{i=1}^G \hat{L}_i(\theta)
\right]
- \beta\, D_{\mathrm{KL}}(\pi_\theta \,\|\, \pi_{\mathrm{ref}})
\end{aligned}
\right.
\end{equation}

where $\mathbb{D}_{KL}$ is the KL divergence regularization term to prevent the policy from deviating too far from the reference model $\pi_{ref}$.
In our experiments we primarily adopt GRPO for its simplicity and stability, but the same SGVR reward can also be optimized with standard PPO~\cite{schulman2017proximal}, as explored in the ablation studies (Section~\ref{sec:ablation}).

\section{Experiments}
\label{sec:exp}

\begin{figure*}[!t]
\centering
\includegraphics[width=0.98\textwidth]{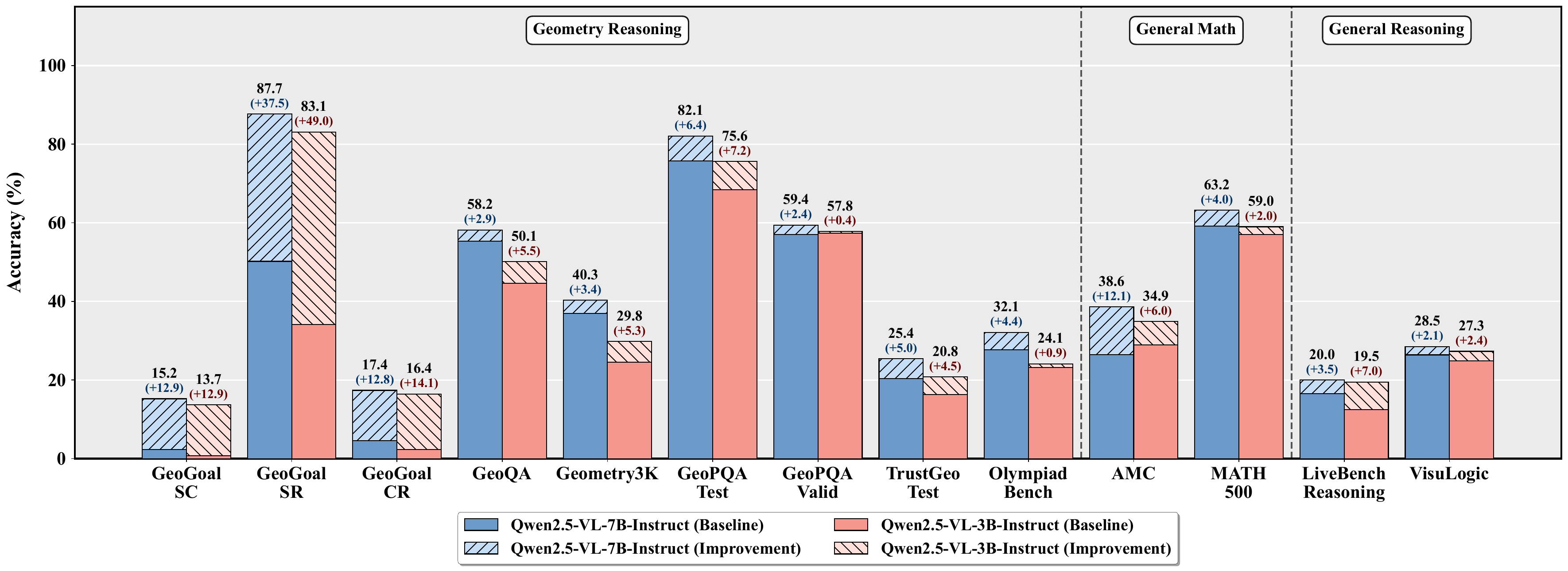}
\caption{Performance comparison of our trained models against baselines on final answer accuracy. Solid bars represent baseline performance; patterned sections indicate improvements from our training. Our method achieves consistent gains across model sizes and task domains, with particularly strong improvements on some datasets (GeoPQA-Test: +6.4\% for 7B, +7.2\% for 3B; AMC: +12.1\% for 7B, +6.0\% for 3B; LiveBench-Reasoning: +3.5\% for 7B, +7.0\% for 3B).}
\label{fig:main_results}
\end{figure*}

\subsection{Experimental Setup}

\paragraph{Training Setup.}
We train Qwen2.5-VL-7B-Instruct and Qwen2.5-VL-3B-Instruct~\cite{bai2025qwen2} using our proposed \textit{SGVR} algorithm on the training split of \textit{GeoGoal} benchmark. This training set consists of 256 plane geometry problems, each with step-wise intermediate sub-goals and final answers to enabling precise subgoal supervision and training details are provided in Appendix~\ref{appendix:training_configs}.

\paragraph{Evaluation Benchmarks.}
To assess distributional robustness and cross-domain generalization, we evaluate our models on benchmarks across three categories: \textbf{\textit{1) Geometric Reasoning:}} We evaluate plane geometry problem-solving capabilities using GeoGoal, GeoQA\citep{chen2025geopqa}, Geometry3K\citep{lu2021inter}, GeoPQA\citep{chen2025geopqa}, TrustGeo-Test\citep{trustgeogen2025}, and OlympiadBench-Geo~\citep{he2024olympiadbench}. These datasets provide a diverse testbed for multimodal geometric reasoning across varying distributions. \textbf{\textit{2) General Mathematics:}} We employ AMC\footnote{\url{https://huggingface.co/datasets/AI-MO/aimo-validation-amc}} and MATH-500~\citep{lightman2023let} to examine whether subgoal-level supervision transfers from geometry to broader mathematical problem-solving. \textbf{\textit{3) General Reasoning:}} We use LiveBench-Reasoning\citep{white2024livebench} and VisuLogic\citep{xu2025visulogic} to probe reasoning capabilities in wider logical and visual contexts.





\paragraph{Evaluation Metric.}

We report \textit{Final Answer Accuracy} as the primary metric across all benchmarks to assess end-to-end reasoning performance. Numerical equivalence is verified using LLM as a deterministic checker for mathematical expressions (more details are provided in Appendix~\ref{app:eval_prompt}). For our \textit{GeoGoal} benchmark, we additionally employ \textit{SC, SR} and \textit{CR} to explicitly evaluate the correctness of intermediate sub-goals. For external benchmarks lacking ground-truth sub-goals, we adopt the established \textit{Process Evaluation Score}~\cite{zhang2025deeptheorem} via an LLM-as-a-Judge approach to assess the quality of intermediate reasoning steps.



\subsection{Main Results}

We structure our analysis around three key research questions to assess the impact of verifiable sub-goal supervision on geometric performance, cross-domain generalization, and reasoning quality.

\vspace{0.5em}
\noindent\textbf{\textit{RQ1: Does rewarding verifiable sub-goals improve geometric reasoning?}}

\noindent Figure~\ref{fig:main_results} presents a comprehensive evaluation across geometry benchmarks, demonstrating that our SGVR framework consistently enhances performance compared to pretrained baselines. 
On our \textit{GeoGoal} benchmark, we observe substantial improvements across all subgoal-level metrics: the 7B model improves from 50.2\% to 87.7\% in Skeleton Rate (SR), from 2.3\% to 15.2\% in Skeleton Completion (SC), and from 4.6\% to 17.4\% in Consistency Ratio (CR); the 3B model shows even more dramatic gains, improving from 34.1\% to 83.1\% in SR, from 0.8\% to 13.7\% in SC, and from 2.3\% to 16.4\% in CR. These improvements validate that our verifiable sub-goal supervision effectively guides models toward more reliable reasoning chains.
On \textbf{external geometry benchmarks}, the 7B model achieves an average accuracy gain of 4.0\%, with notable improvements on GeoPQA-Test (+6.4\%) and TrustGeo-Test (+5.0\%).
The 3B model mirrors this trend with GeoPQA-Test (+7.2\%) and Geometry3K (+5.3\%). These results demonstrate that verifiable sub-goal supervision facilitates robust generalization within the geometric domain.

\vspace{0.5em}
\noindent\textbf{\textit{RQ2: Do geometric sub-goal priors generalize to non-geometric domains?}}

\noindent A critical question is whether the reasoning capabilities learned from geometry are specific to that domain or transferable to broader contexts. Despite being trained exclusively on geometry-focused data without exposure to general math or logic samples, our models exhibit remarkable plasticity and cross-domain generalization. 
In general mathematics, the models demonstrate significant performance boosts, with the 7B model improving by 12.1\% on the AMC benchmark and 4.0\% on MATH-500, while the 3B model shows respective gains of 6.0\% and 2.0\%. This indicates that the verification mechanism learned from geometric sub-goals effectively supports broader symbolic mathematical reasoning.
Moreover, these benefits extend to general reasoning tasks, as evidenced by the 7B and 3B models achieving gains of 3.5\% and 7.0\% respectively on LiveBench-Reasoning, alongside consistent improvements on the visual logic benchmark VisuLogic. This suggests that the rigorous verification of geometric sub-goals cultivates a fundamental reasoning capability that naturally transfers to enhance logical consistency across diverse domains.

\begin{table}[t]
\centering
\small
\caption{\textit{Process evaluation scores}~\cite{zhang2025deeptheorem} across all benchmarks. Models trained with our method consistently improve the quality of reasoning process over the pretrained baselines across both model sizes.}
\label{tab:process_eval}
\begin{adjustbox}{width=0.78\linewidth}
\begin{tabular}{lcccccc}
\toprule
\multirow{2}{*}{\textbf{Benchmark}}  & \multicolumn{3}{c}{\textbf{Qwen2.5-VL-7B}} & \multicolumn{3}{c}{\textbf{Qwen2.5-VL-3B}} \\
\cmidrule(lr){2-4} \cmidrule(lr){5-7}
& Baseline & Ours & $\Delta$ & Baseline & Ours & $\Delta$ \\
\midrule
GeoGoal & 15.7 & \textbf{23.7} & \cellcolor{green!15}+8.0 & 13.0 & \textbf{26.9} & \cellcolor{green!15}+13.9 \\
GeoQA & 49.0 & \textbf{50.8} & \cellcolor{green!15}+1.8 & 40.4 & \textbf{42.6} & \cellcolor{green!15}+2.2 \\
Geometry3K & 50.5 & \textbf{52.0} & \cellcolor{green!15}+1.5 & 41.3 & \textbf{43.9} & \cellcolor{green!15}+2.6 \\
GeoPQA-Test & 47.2 & \textbf{63.6} & \cellcolor{green!15}+16.4 & 23.3 & \textbf{46.1} & \cellcolor{green!15}+22.8 \\
GeoPQA-Valid & 28.9 & \textbf{35.9} & \cellcolor{green!15}+7.0 & 29.1 & \textbf{35.1} & \cellcolor{green!15}+6.0 \\
TrustGeo-Test & 23.9 & \textbf{26.1} & \cellcolor{green!15}+2.2 & 17.8 & \textbf{20.2} & \cellcolor{green!15}+2.4 \\
OlympiadBench & 36.7 & \textbf{39.9} & \cellcolor{green!15}+3.2 & 29.2 & \textbf{30.7} & \cellcolor{green!15}+1.5 \\
AMC & 38.0 & \textbf{40.3} & \cellcolor{green!15}+2.3 & 35.8 & \textbf{40.2} & \cellcolor{green!15}+4.4 \\
MATH 500 & 66.8 & \textbf{69.0} & \cellcolor{green!15}+2.2 & 65.6 & \textbf{66.5} & \cellcolor{green!15}+0.9 \\
LiveBench-Reasoning & 20.3 & \textbf{22.0} & \cellcolor{green!15}+1.7 & 17.4 & \textbf{20.5} & \cellcolor{green!15}+3.1 \\
VisuLogic & 34.7 & \textbf{35.3} & \cellcolor{green!15}+0.6 & 31.1 & 31.1 & \cellcolor{gray!15}+0.0 \\
\bottomrule
\end{tabular}
\end{adjustbox}
\end{table}

\vspace{0.5em}
\noindent\textbf{\textit{RQ3: Does sub-goal alignment improve the quality of the reasoning chain?}}

\noindent To assess reasoning fidelity beyond final outcomes, we evaluate the logical coherence of generated paths using the established \textit{Process Evaluation Score}~\cite{zhang2025deeptheorem}.
Table~\ref{tab:process_eval} reveals a universal improvement in process scores across nearly all evaluated benchmarks for both model sizes, indicating a broad enhancement in reasoning quality.
On \textit{GeoGoal}, process scores improve from 15.7\% to 23.7\% (+8.0\%) for the 7B model and from 13.0\% to 26.9\% (+13.9\%) for the 3B model, demonstrating that sub-goal alignment significantly enhances the quality of intermediate reasoning steps.
On external benchmarks, the GeoPQA-Test process score increases by 16.4\% for the 7B model and by an impressive 22.8\% for the 3B model.
These findings provide compelling evidence that SGVR encourages the generation of more reliable and coherent intermediate trajectories.
\begin{table}[t]
\centering
\small
\caption{Ablation study of RL optimizers (Baseline, PPO, GRPO) on Qwen2.5-VL-7B-Instruct and Qwen2.5-VL-3B-Instruct across benchmarks. Both algorithms use Skeleton Rate as the reward signal. Best results within each model size are in \textbf{bold}, second-best are \underline{underlined}.}
\label{tab:rl_algorithm}
\begin{adjustbox}{width=0.78\linewidth}
\begin{tabular}{lcccccc}
\toprule
\multirow{2}{*}{\textbf{Benchmark}} & \multicolumn{3}{c}{\textbf{Qwen2.5-VL-7B}} & \multicolumn{3}{c}{\textbf{Qwen2.5-VL-3B}} \\
\cmidrule(lr){2-4} \cmidrule(lr){5-7}
& Baseline & PPO & GRPO & Baseline & PPO & GRPO \\
\midrule
\multicolumn{7}{l}{\textbf{\textit{Geometry Reasoning}}} \\
\midrule
GeoGoal (SC) & 2.3 & \underline{10.6} & \textbf{15.2} & 0.8 & \underline{11.7} & \textbf{13.7} \\
GeoGoal (SR) & 50.2 & \underline{80.8} & \textbf{87.7} & 34.1 & \textbf{84.0} & \underline{83.1} \\
GeoGoal (CR) & 4.6 & \underline{13.1} & \textbf{17.4} & 2.3 & \underline{13.9} & \textbf{16.4} \\
GeoQA & 55.3 & \underline{56.2} & \textbf{58.2} & 44.6 & \textbf{50.1} & \textbf{50.1} \\
Geometry3K & 36.9 & \underline{39.1} & \textbf{40.3} & 24.5 & \textbf{29.8} & \textbf{29.8} \\
GeoPQA-Test & 75.7 & \underline{78.8} & \textbf{82.1} & 68.4 & \textbf{78.6} & \underline{75.6} \\
GeoPQA-Valid & 57.0 & \textbf{64.6} & \underline{59.4} & 57.4 & \textbf{60.3} & \underline{57.8} \\
TrustGeo-Test & 20.4 & \underline{21.7} & \textbf{25.4} & 16.3 & \underline{20.4} & \textbf{20.8} \\
OlympiadBench & 27.7 & \underline{29.5} & \textbf{32.1} & 23.2 & \textbf{25.9} & \underline{24.1} \\
\textit{Geometry Average} & 36.7 & \underline{43.8} & \textbf{46.4} & 30.2 & \textbf{41.6} & \underline{41.3} \\
\midrule
\multicolumn{7}{l}{\textbf{\textit{General Mathematics}}} \\
\midrule
AMC & 26.5 & \underline{33.7} & \textbf{38.6} & 28.9 & \textbf{34.9} & \textbf{34.9} \\
MATH 500 & 59.2 & \underline{62.6} & \textbf{63.2} & 57.0 & \textbf{59.2} & \underline{59.0} \\
\textit{General Math Average} & 42.9 & \underline{48.2} & \textbf{50.9} & 43.0 & \textbf{47.1} & \underline{47.0} \\
\midrule
\multicolumn{7}{l}{\textbf{\textit{General Reasoning}}} \\
\midrule
LiveBench-Reasoning & 16.5 & \textbf{21.0} & \underline{20.0} & 12.5 & \textbf{20.5} & \underline{19.5} \\
VisuLogic & 26.4 & \textbf{28.6} & \underline{28.5} & 24.9 & \underline{25.9} & \textbf{27.3} \\
\textit{General Reasoning Average} & 21.5 & \textbf{24.8} & \underline{24.3} & 18.7 & \underline{23.2} & \textbf{23.4} \\
\midrule
\textbf{Overall Average} & 33.7 & \underline{39.0} & \textbf{40.5} & 30.6 & \textbf{37.3} & \underline{37.2} \\
\bottomrule
\end{tabular}
\end{adjustbox}
\end{table}
\begin{table}[t]
\centering
\small
\caption{Ablation study of reward formulations (Final Answer, Skeleton Completion, Skeleton Rate) on Qwen2.5-VL-7B-Instruct and Qwen2.5-VL-3B-Instruct across benchmarks in different domains.}
\label{tab:ablation}
\begin{adjustbox}{width=0.78\linewidth}
\begin{tabular}{lcccccc}
\toprule
\multirow{2}{*}{\textbf{Benchmark}} & \multicolumn{3}{c}{\textbf{Qwen2.5-VL-7B}} & \multicolumn{3}{c}{\textbf{Qwen2.5-VL-3B}} \\
\cmidrule(lr){2-4} \cmidrule(lr){5-7}
& \textbf{FA} & \textbf{SC} & \textbf{SR} & \textbf{FA} & \textbf{SC} & \textbf{SR} \\
& \textbf{Reward} & \textbf{Reward} & \textbf{Reward} & \textbf{Reward} & \textbf{Reward} & \textbf{Reward} \\
\midrule
\multicolumn{7}{l}{\textbf{\textit{Geometry Reasoning}}} \\
\midrule
GeoGoal (SC) & 5.9 & \underline{9.4} & \textbf{15.2} & 5.9 & \underline{9.4} & \textbf{13.7} \\
GeoGoal (SR) & \underline{79.1} & 82.1 & \textbf{87.7} & \underline{79.1} & 67.5 & \textbf{83.1} \\
GeoGoal (CR) & \underline{7.4} & 11.4 & \textbf{17.4} & \underline{7.4} & 13.9 & \textbf{16.4} \\
GeoQA & \underline{54.4} & 53.9 & \textbf{58.2} & 48.0 & \underline{48.3} & \textbf{50.1} \\
Geometry3K & 36.7 & \underline{36.8} & \textbf{40.3} & \underline{27.0} & 26.0 & \textbf{29.8} \\
GeoPQA-Test & \underline{77.9} & 77.5 & \textbf{82.1} & 72.2 & \underline{74.2} & \textbf{75.6} \\
GeoPQA-Valid & \underline{42.6} & 40.8 & \textbf{59.4} & \underline{57.6} & 57.4 & \textbf{57.8} \\
TrustGeo-Test & \underline{20.8} & 19.6 & \textbf{25.4} & \underline{20.0} & 13.8 & \textbf{20.8} \\
OlympiadBench & \underline{27.7} & 24.1 & \textbf{32.1} & \underline{21.4} & 21.4 & \textbf{24.1} \\
\textit{Geometry Average} & 39.2 & \underline{39.5} & \textbf{46.4} & \underline{37.6} & 36.9 & \textbf{41.3} \\
\midrule
\multicolumn{7}{l}{\textbf{\textit{General Mathematics}}} \\
\midrule
AMC & 31.3 & \underline{33.7} & \textbf{38.6} & \underline{31.3} & 28.9 & \textbf{34.9} \\
MATH 500 & 61.2 & \textbf{65.6} & \underline{63.2} & \underline{56.6} & 54.6 & \textbf{59.0} \\
\textit{General Math Average} & 46.3 & \underline{49.7} & \textbf{50.9} & \underline{44.0} & 41.8 & \textbf{47.0} \\
\midrule
\multicolumn{7}{l}{\textbf{\textit{General Reasoning}}} \\
\midrule
LiveBench-Reasoning & 17.5 & \textbf{25.0} & \underline{20.0} & \underline{17.5} & 13.5 & \textbf{19.5} \\
VisuLogic & 25.6 & \underline{26.5} & \textbf{28.5} & \underline{25.1} & 25.0 & \textbf{27.3} \\
\textit{General Reasoning Average} & 21.6 & \textbf{25.8} & \underline{24.3} & \underline{21.3} & 19.3 & \textbf{23.4} \\
\midrule
\textbf{Overall Average} & 35.7 & \underline{38.3} & \textbf{40.5} & \underline{34.3} & 32.7 & \textbf{37.2} \\
\bottomrule
\end{tabular}
\end{adjustbox}
\end{table}

\subsection{Ablation Studies}
\label{sec:ablation}
To identify the optimal configuration for our SGVR framework, we conduct ablation studies focusing on two critical design choices: the choice of reinforcement learning optimizer and the granularity of the verifiable reward signal.

\vspace{0.5em}
\noindent\textit{\textbf{RQ4: Is the performance gain sensitive to the choice of RL optimizer?}}

\noindent To evaluate the impact of different optimization strategies within our framework, we compare Group Relative Policy Optimization (GRPO)~\citep{deepseek2025deepseek} against standard Proximal Policy Optimization (PPO)~\citep{schulman2017proximal}, both utilizing identical Skeleton Rate rewards. 
Our analysis of Table~\ref{tab:rl_algorithm} yields three key observations regarding optimizer efficacy: 

\noindent\textit{(1) Both optimizers consistently improve over the pretrained baseline.} Across almost all benchmarks and both model scales, models trained with PPO and GRPO outperform the baseline. 
For the 7B model, PPO and GRPO achieve averages of 39.0\% and 40.5\% respectively among all benchmark, compared to the baseline average of 33.7\%. For the 3B model, they reach 37.3\% and 37.2\% versus 30.6\%. This indicates that SGVR framework is effective and robust under different optimization schemes.

\noindent\textit{(2) GRPO and PPO exhibit complementary strengths for the 7B model.} GRPO achieves the best overall performance of 40.5\%, demonstrating stronger results on mathematical and geometric tasks. For instance, it leads in the Geometry Average with 46.4\% versus 43.8\% for PPO, and in the General Math Average with 50.9\% versus 48.2\%. Conversely, PPO performs better on general reasoning tasks, achieving an average of 24.8\% compared to 24.3\% for GRPO.


\vspace{0.5em}
\noindent{\textbf{\textit{RQ5: Which reward formulation maximizes reasoning performance?}}}

\noindent To determine the optimal supervision signal, we compare three reward formulations defined in Section~\ref{sec:eval_metrics}: \textbf{1) Skeleton Rate} which provides dense sub-goal rewards, \textbf{2) Skeleton Completion} which enforces strict sub-goal reward, and \textbf{3) Final Answer} which relies on sparse outcome signals. 
Our analysis on Table~\ref{tab:ablation} leads to three key observations:

\noindent\textit{(1) Skeleton Rate offers the most robust and effective supervision.} SR consistently achieves the highest overall average performance, reaching 40.5\% (7B) and 37.2\% (3B). On \textit{GeoGoal}, SR demonstrates superior reasoning integrity, boosting SC scores to 15.2\% (7B) and 13.7\% (3B), and CR scores to 17.4\% (7B) and 16.4\% (3B). This superiority extends beyond geometry, with SR also achieving the highest averages in general mathematics and competitive results in general reasoning.

\noindent{\textit{(2) Sparse Final Answer rewards are insufficiently informative for complex reasoning.}} 
Outcome-based supervision fails to foster consistent reasoning chains. On the \textit{GeoGoal} benchmark, models trained with FA rewards show a sharp disconnect between subgoal-wise correctness and full-chain validity: despite achieving high SR scores (e.g., 79.1\% for 7B), their SC and CR metrics collapse to just 5.9\% and 7.4\%, respectively. This pattern holds across domains, where FA training consistently results in the lowest average performance for both model sizes, trailing SR by significant margins in geometry (−7.2\%), general mathematics (−4.6\%) and general reasoning (-2.7\%) for the 7B model.



We therefore adopt \textbf{\textit{Skeleton Rate (SR)}} as our default reward formulation, as its dense reward offers the most stable and effective signal for fostering robust reasoning capabilities.

\begin{figure}[t]
\centering
\includegraphics[width=0.78\textwidth]{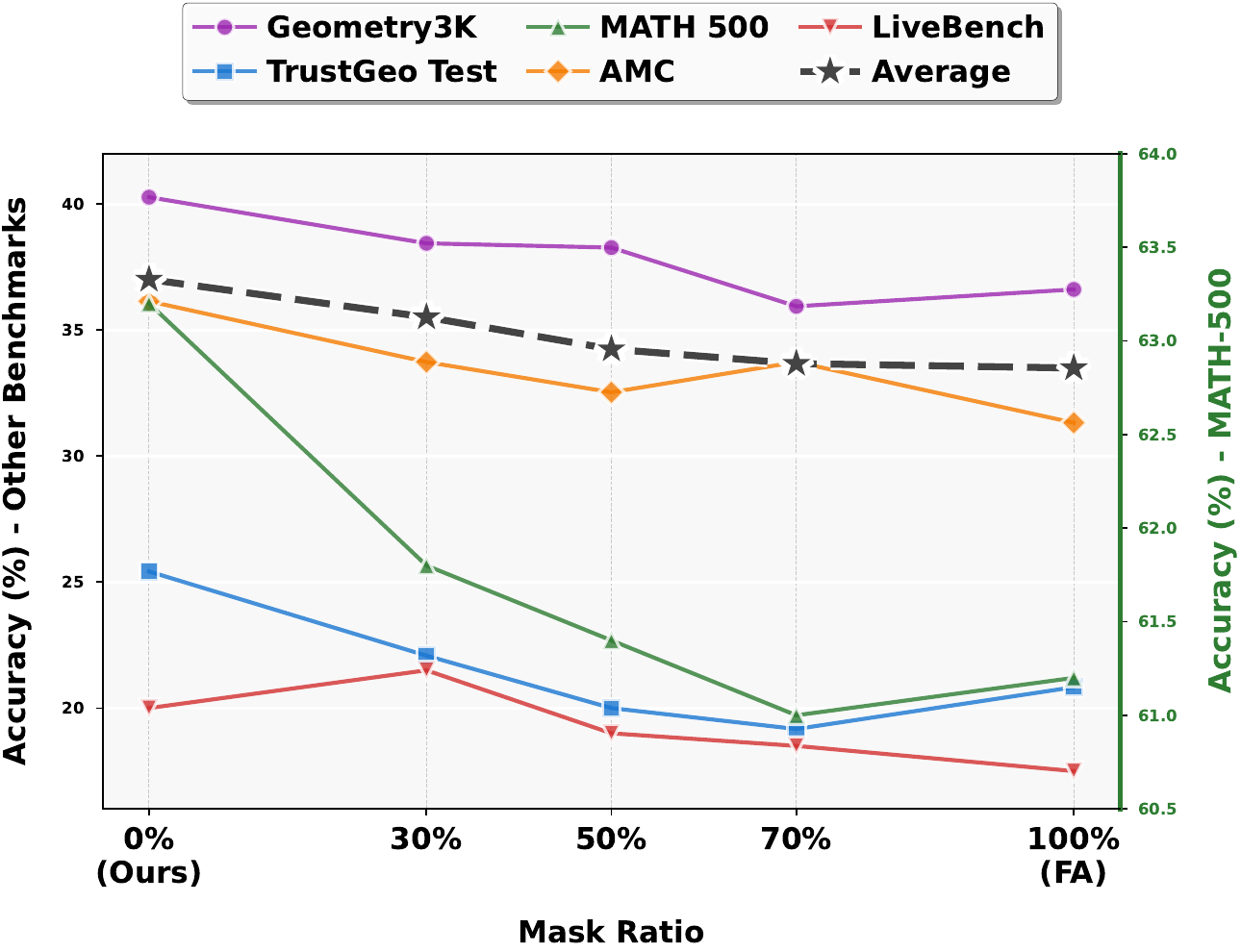}
\caption{Performance comparison across different sub-goal mask ratios trained on Qwen2.5-VL-7B-Instruct.}

\label{fig:mask_ablation}
\end{figure}

\vspace{0.5em}
\noindent{\textbf{\textit{RQ6: Is denser sub-goal supervision always better?}}}

\noindent To investigate whether more subgoals always lead to better performance, we conduct an ablation study by randomly masking a proportion of subgoals during training while maintaining the Skeleton Rate reward formulation. We train models with 0\%, 30\%, 50\%, 70\%, and 100\% (Final Answer only) of subgoals masked.
Figure~\ref{fig:mask_ablation} presents the results across five benchmarks, revealing two key observations:

\noindent\textit{(1) Denser supervision generally improves performance.} The 0\% mask ratio achieves the highest accuracy on most benchmarks, with the overall average accuracy decreasing consistently from 37.01\% (0\%) to 33.49\% (100\%). This indicates that additional sub-goals provide valuable supervision signals for learning robust reasoning strategies.

\noindent\textit{(2) Optimal sub-goal density for generalization is task-specific.} Our results show that while in-domain geometric tasks demand fine-grained steps to maintain logical rigor, out-of-domain generalization can sometimes favor sparser signals. For instance, on LiveBench-Reasoning, the 30\% configuration achieves the highest accuracy. This suggests that transferring to broader domains may benefit from focusing on key milestones rather than pursuing the densest possible in-domain signals.

\vspace{0.3em}
\section{Conclusion}
\label{sec:conclusion}

In this work, we introduce a paradigm shift from outcome-based to subgoal-level supervision.
We construct \textit{GeoGoal} via a formal verification engine to provide verifiable numeric sub-goals and propose the \textit{SGVR} framework to leverage these as dense reward signals.
Our approach significantly enhances in-domain geometric reasoning while demonstrating strong transferability to general mathematics and broader reasoning tasks.
Crucially, our findings suggest that developing post-training methods within in-domain formal engines capable of providing trustworthy dense signals offers a promising avenue for unlocking robust out-of-distribution generalization capabilities.

\section{Limitations and Future Work}
Our benchmark is derived from a specific formal data engine and a mapping into numeric sub-goals, which may not capture the full diversity of human-written geometric arguments or non-numeric intermediate reasoning.
Moreover, verification relies on deterministic equivalence checking for numeric answers; extending verification to richer symbolic and diagram-grounded statements remains an open challenge.

Future work includes adopting more general-purpose formal systems (e.g., Lean4)~\cite{moura2021lean} to extend beyond geometry into broader mathematical domains, and transferring the SGVR decomposition-and-verification paradigm to other reasoning tasks where intermediate sub-goals can be designed to be automatically checkable.
\clearpage
\bibliography{iclr2026_conference}
\bibliographystyle{iclr2026_conference}

\clearpage
\appendix

\section{Ethical Considerations}
The GeoGoal benchmark is synthesized using the formal verification engine, TrustGeoGen~\cite{trustgeogen2025}, which generates problems and reasoning paths based on rigorous axiomatic systems rather than crawling private or copyrighted web content. Furthermore, we utilize established open-source datasets (e.g., MATH, LiveBench)~\cite{lightman2023let,white2024livebench} in strict accordance with their original licensing terms. All data used in this study is intended for academic research and contains no personally identifiable information (PII) or harmful content.

\section{Related Work}

\subsection{Multimodal LLMs and Visual Mathematical Reasoning}

Large language models (LLMs) have achieved remarkable progress in linguistic intelligence across a wide spectrum of tasks~\citep{chatgpt/ouyang2022training,llama/touvron2023llama,llama2/touvron2023llama,internlm/team2023internlm}. Building on this foundation, multimodal large language models (MLLMs) incorporate visual processing capabilities via modality-alignment modules, such as Q-Former~\citep{blip-2/li2023blip} and lightweight projection layers~\citep{llava/liu2024visual}. These architectures have demonstrated strong performance on general vision-language benchmarks~\citep{fu2023mme,xia2024docgenome,structchart/xia2023structchart,mathvista/lumathvista,jiang2025mme}.

However, a critical "visual-reasoning gap" persists despite these perceptual gains. Recent studies indicate that MLLMs frequently suffer from \textit{object hallucination}~\citep{li2023evaluating} and struggle to maintain logical consistency between visual perception and textual deduction~\citep{liu2023hallusionbench}. In the geometric domain, this issue is particularly acute, manifesting as the reasoning illusion''~\citep{Wang2025DoLL} where models may retrieve correct formulas but apply them to hallucinated geometric primitives. As emerging systems increasingly position MLLMs as scientific agents interacting with complex environments~\citep{team2025novelseek,zhang2025origene,gottweis2025towards}, the demand for rigorous reasoning has intensified. To address this, Multimodal Chain-of-Thought strategies~\citep{zhang2023multimodal,shao2023prompting} have been proposed to bridge the gap between visual perception and answer generation by eliciting intermediate rationales.

Nevertheless, when confronted with visual mathematical content such as geometry diagrams, current MLLMs continue to exhibit significant performance drops. This is largely attributed to the domain discrepancy between natural images and schematic figures, as well as the requirement for long-horizon, logically precise reasoning~\citep{mathvista/lumathvista}. To mitigate these limitations, domain-specialized models have leveraged targeted data or training objectives: MAVIS synthesizes large-scale chain-of-thought supervision for math diagrams~\citep{mavis/zhang2024mavis}, while G-LLaVA collects supermodel-guided geometric solutions~\citep{g-llava/gao2023g}. Similarly, GeoX aligns visual features with formal geometric primitives to enable solver-backed theorem verification~\citep{geox/Xia2024GeoXGP}. Our work offers a complementary perspective: rather than proposing a new architecture, we focus on extracting verifiable process signals from formal geometric structures and utilizing them as dense rewards to enhance the reasoning reliability of existing MLLMs.

\subsection{Geometric Problem Solving with MLLMs and Formal Solvers}

Automatic geometric problem solving (GPS) requires understanding diagrams, interpreting symbolic conditions, and composing nontrivial deductive chains. A line of work enhances visual and textual understanding via unimodal pre-training, cross-modal alignment, and instruction tuning on geometry corpora~\citep{geoqa_ngs/chen2021geoqa,unigeo_geoformer/chen2022unigeo,pgpsnet/zhang2023multi,pgdp5k/hao2022pgdp5k,geox/Xia2024GeoXGP,mavis/zhang2024mavis,g-llava/gao2023g,jiang2025mme}. These methods typically train MLLMs to directly output final numerical answers or natural-language solutions given the diagram and problem text.

Another line of research adopts formal geometric solvers or external interpreters. Systems such as AlphaGeometry and its successors~\citep{trinh2024solving,alphageometry2/chervonyi2025,sicca2024newclid} can solve problems at the level of international mathematical olympiads by encoding each instance in a formal language and searching in a rule-based state space. While these approaches offer strong guarantees and IMO-level performance, they require precise symbolic modeling of each instance, which limits their practicality for open-ended user-facing applications. To bridge the gap between rigorous calculation and open-ended reasoning, Program-of-Thought~\citep{chen2022program} and PAL~\citep{gao2023pal} decouple reasoning from computation by delegating arithmetic to external Python interpreters. While effective for reducing calculation errors, these methods largely treat reasoning as a linear script generation task without verifying the logical soundness of the underlying deductive chain.

Our work lies in between: we rely on a formal geometric backend to generate trusted reasoning skeletons, but keep the inference model as an MLLM that operates directly over diagrams and text. Instead of asking the solver to produce complete symbolic proofs at test time, we convert its offline skeletons into verifiable subgoals and use them to shape the MLLM's reasoning process through reinforcement learning.

\subsection{Datasets and Benchmarks for Geometric Reasoning}

High-quality data is critical for improving GPS systems. Existing datasets can be roughly divided into three construction paradigms~\citep{geoqa_ngs/chen2021geoqa,dpe-gps/cao2022augmented,unigeo_geoformer/chen2022unigeo,mathvista/lumathvista,he2024olympiadbench}. The first filters real-world exam or textbook problems and manually annotates diagrams and solutions, as in GeoQA~\citep{geoqa_ngs/chen2021geoqa}, GeoQA+~\citep{dpe-gps/cao2022augmented}, UniGeo~\citep{unigeo_geoformer/chen2022unigeo}, PGDP5K~\citep{pgdp5k/hao2022pgdp5k}, MathVista~\citep{mathvista/lumathvista}, and OlympiadBench~\citep{he2024olympiadbench}. These datasets offer human-authored, high-quality questions, but their scalability is constrained by limited source pools and annotation cost, and the difficulty level is often biased toward middle- or high-school geometry.

In contrast to manual annotation, the second paradigm uses formal engines to synthesize problems and proofs~\citep{lu2021inter,pgpsnet/zhang2023multi,Kazemi2023GeomVerseAS}. Inter-GPS and PGPS9K generate diagram–text pairs by sampling from pre-defined geometry configurations~\citep{lu2021inter,pgpsnet/zhang2023multi}, while GeomVerse augments authentic questions via LLM-based transformations~\citep{Kazemi2023GeomVerseAS}. Formal engines can guarantee logical correctness and scale up easily, but the resulting textual solutions may diverge from natural mathematical discourse. The third paradigm employs LLMs to synthesize reasoning trajectories~\citep{mavis/zhang2024mavis,g-llava/gao2023g}, which yields human-like step-by-step explanations but lacks verifiable guarantees and may introduce subtle logical errors. Recent studies suggest that high-quality synthetic data is crucial for unlocking complex reasoning capabilities~\citep{li2023textbooks,gunasekar2023textbooks}. However, synthesizing reliable geometric reasoning paths remains challenging due to the difficulty of ensuring cross-modal consistency between diagrams and text.

More recently, TrustGeoGen~\citep{trustgeogen2025} proposes a scalable, rule-driven engine that generates synthetic geometry problems together with formal proofs, natural-language explanations, and diagrams under a unified formal language. GeoBench~\citep{feng2025geobenchrethinkingmultimodalgeometric} further builds on TrustGeoGen to design a hierarchical GPS benchmark that evaluates four critical abilities: visual perception, goal-oriented planning, rigorous theorem application, and self-reflective backtracking, moving beyond single final-answer accuracy. Other works such as GeomRel~\citep{Wang2025DoLL} and GeoSense~\citep{Xu2025GeoSenseEI} explore structural diagram understanding and theorem-application patterns but still focus on narrow subskills. In contrast, our work leverages formal skeletons from a TrustGeoGen-style engine to construct a sequence of verifiable numeric subgoals for each instance and defines skeleton-based metrics (Skeleton Rate and Skeleton Completion) that jointly capture local step correctness and global proof coherence. Importantly, we go one step further by using these verifiable subgoals not only for evaluation but also as dense training signals for reinforcement learning.

\subsection{Process Supervision and Reinforcement Learning for Reasoning}

There is a growing interest in process-oriented supervision for mathematical and logical reasoning. Recent benchmarks and evaluators~\citep{mathvista/lumathvista,zhang2025deeptheorem} analyze the quality of intermediate steps rather than only final answers, revealing phenomena such as ``shortcut'' solutions and self-contradictory chains. In geometric reasoning, GeoBench~\citep{feng2025geobenchrethinkingmultimodalgeometric} evaluates models at multiple levels (from perception to backtracking) using structured tasks derived from formal reasoning graphs, but the resulting signals are used purely for diagnosis. 

While Process Reward Models~\citep{lightman2023let,uesato2022solving} have successfully scaled mathematical reasoning by training discriminator models to score intermediate steps, they face a fundamental bottleneck: the reliance on expensive human annotations or synthesized labels. Furthermore, learned reward models are susceptible to ``reward hacking,'' where the policy model learns to exploit the critic's inaccuracies rather than improving reasoning quality~\citep{gao2023scaling}. In parallel, reasoning-optimized models such as OpenAI o1/o3~\citep{openai-o1,openai-o3} and specialized MLLMs~\citep{Wu2024DeepSeekVL2MV,gemini-2.5-pro,bai2025qwen2} implicitly incorporate internal process supervision and reinforcement learning, but their training recipes and reward functions are largely proprietary or rely on learned reward models that can themselves be unreliable. 

Our work is closest in spirit to process-supervised and RL-based reasoning, but differs in two key aspects. First, we obtain \emph{rule-grounded} milestone signals by decomposing formal proof skeletons into atomic sub-goals and mapping them to numeric targets that can be automatically verified for each step. Second, we transform these verifiable sub-goals into token-level advantages and optimize MLLMs with a PPO-style objective, thereby turning skeleton-based evaluation into a dense, stable reward signal. This avoids training a separate reward model and mitigates hallucination in the supervision channel. Experiments across geometric, mathematical, and general reasoning benchmarks show that such verifiable sub-goal rewards not only improve final-answer accuracy but also significantly enhance the quality and consistency of the generated reasoning chains.

\begin{figure*}[t]
  \centering
  \begin{minipage}[b]{0.48\textwidth}
    \centering
    \includegraphics[width=\textwidth]{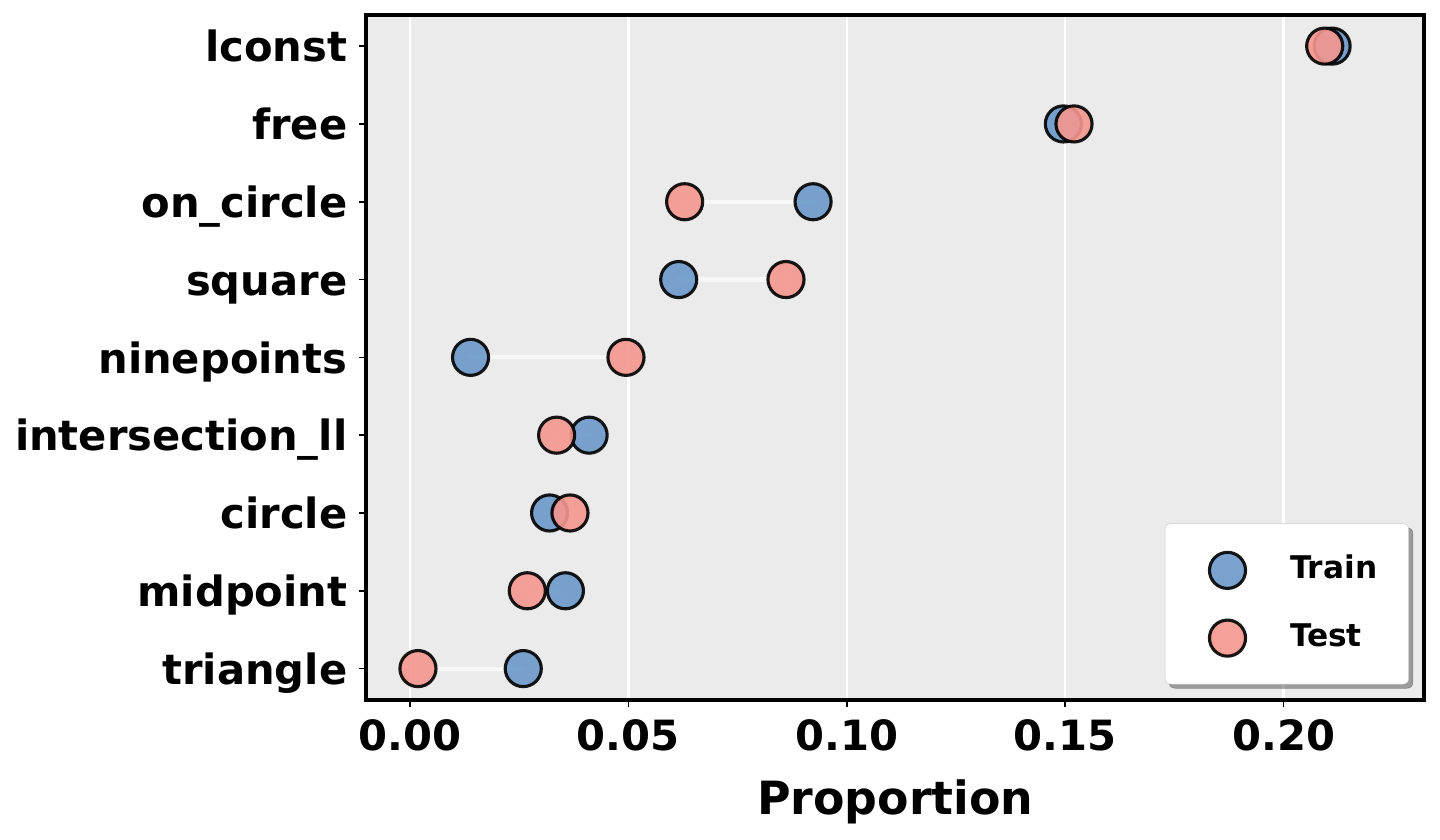}
  \end{minipage}
  \hfill
  \begin{minipage}[b]{0.48\textwidth}
    \centering
    \includegraphics[width=\textwidth]{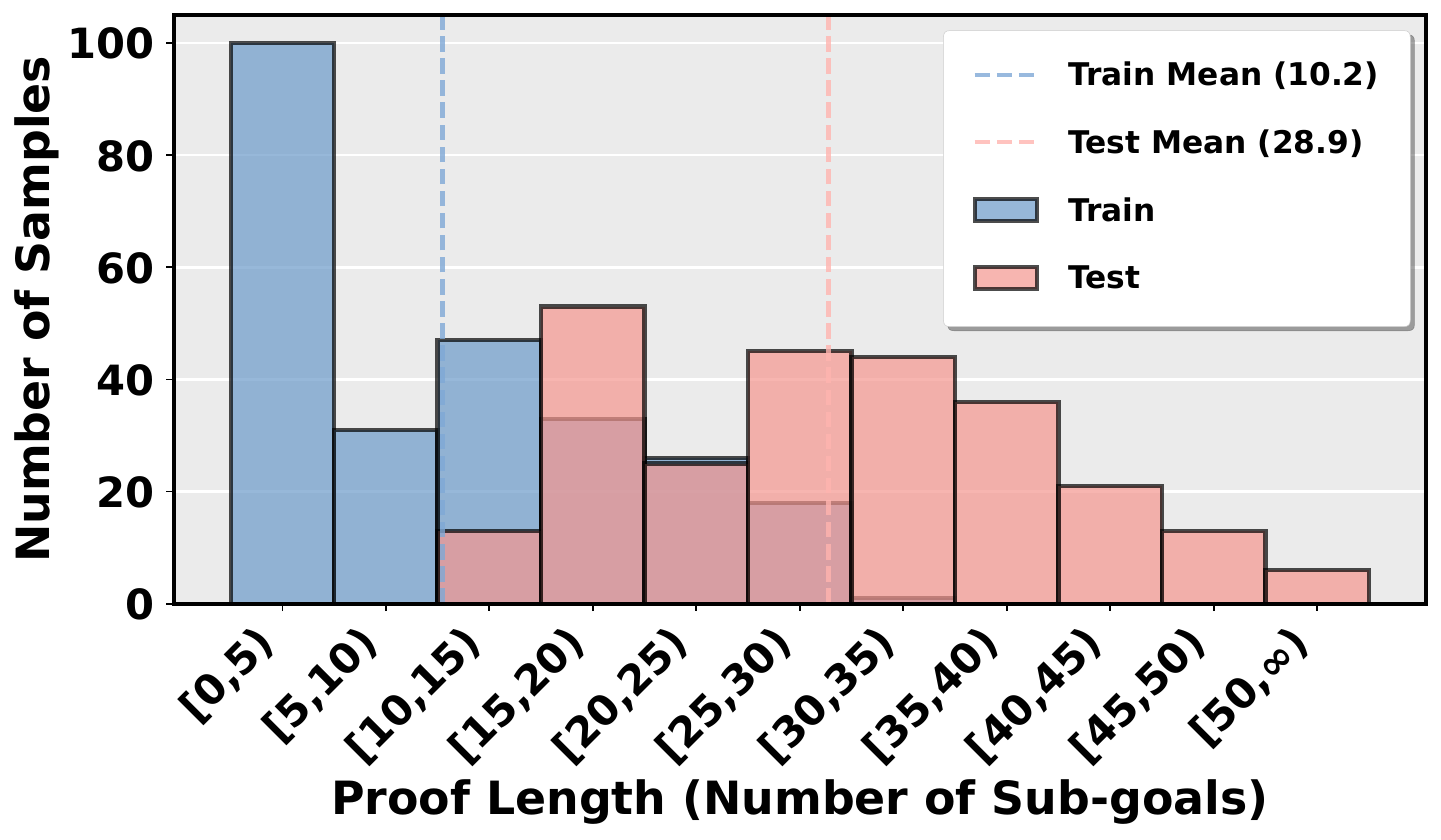}
  \end{minipage}
  \caption{\textbf{Dataset characteristics across Train and Test splits.} Left: Dumbbell chart showing the proportion of geometric concepts in both splits, where connecting lines indicate distributional differences. Right: Histogram showing proof-length distribution, with the test set containing a higher proportion of instances with longer reasoning chains, providing a more challenging evaluation of multi-step reasoning capabilities.}
  \label{fig:dataset_characteristics}
\end{figure*}
\section{Dataset Characteristics}
\label{app:dataset_stats}

This section provides detailed statistics and distributions for our step-wise verifiable geometric reasoning benchmark.

Figure~\ref{fig:dataset_characteristics} presents comprehensive statistics of our benchmark across Train and Test splits. The left panel shows the distribution of geometric concepts (predicates and constructions), where both splits exhibit comparable coverage across all major element types, including constant-valued constraints (\texttt{lconst}), free point constructions (\texttt{free}), circle-related predicates (\texttt{on\_circle}), square constructions, and various intersection and midpoint predicates. The dumbbell chart clearly illustrates where the two splits align (short connecting lines) and where they differ (longer connecting lines). Specialized constructions like centroids, nine-point circles, and parallelograms appear less frequently but are present in both splits. The right panel shows the proof-length distribution, revealing that the test set is intentionally skewed toward longer reasoning chains, with more instances requiring 8+ sub-goal verifications, providing a diverse range of reasoning complexities in our benchmark.

\section{Geometric-to-Numeric Mapping}
\label{app:mapping}

This section specifies the complete mapping from formal geometric predicates to numeric evaluation targets used in our benchmark construction. Each predicate in the formal language (based on the Newclid system~\citep{sicca2024newclid}) is associated with one or more numeric expressions and their corresponding ground-truth values. All angle measurements are expressed in degrees modulo 180°; all ratios are dimensionless.

\subsection{Notation and Conventions}

For each predicate type, we provide:
\begin{itemize}
  \item \textbf{Predicate}: The formal predicate identifier and argument pattern
  \item \textbf{Numeric Form}: The corresponding numeric expression $T$ to be evaluated
  \item \textbf{Expected Value}: The ground-truth value for correct instantiations
  \item \textbf{Notes}: Additional specifications regarding orientation, degenerate configurations, or alternative formulations
\end{itemize}

\noindent\textbf{Notational conventions:}
\begin{itemize}
  \item $|AB|$ denotes the Euclidean length of segment $AB$
  \item $\angle(AB, CD)$ denotes the directed angle between line segments $AB$ and $CD$
  \item $\mathrm{area\_triangle}(A,B,C)$ denotes the signed area of triangle $ABC$
  \item For equality predicates, we adopt ratio-based formulations (expected value 1) rather than difference-based formulations (expected value 0) to mitigate numerical instability near zero
\end{itemize}

\subsection{Core Geometric Predicates}
Table~\ref{tab:mapping_core} specifies the numeric mappings for fundamental geometric predicates that commonly appear in formal proof derivations.

\subsection{Constant Constraints}
Formal proof derivations frequently involve constant-valued constraints on geometric quantities (lengths, angles, ratios). Table~\ref{tab:mapping_constants} specifies their numeric representations.

\subsection{Special Triangle Types}
Table~\ref{tab:mapping_triangles} defines the numeric verification conditions for predicates characterizing specialized triangle configurations.

\subsection{Quadrilaterals and Polygons}
Table~\ref{tab:mapping_quads} specifies the verification conditions for predicates pertaining to quadrilaterals and higher-order polygons.

\begin{table*}[t!] 
\centering
\footnotesize
\caption{Mapping of core geometric predicates to numeric evaluation targets.}
\label{tab:mapping_core}
\begin{adjustbox}{max width=\linewidth}
\begin{tabular}{llcp{5cm}}
\toprule
\textbf{Predicate} & \textbf{Numeric Form $T$} & \textbf{Expected} & \textbf{Notes} \\
\midrule
\multicolumn{4}{l}{\textit{Equality Predicates}} \\
\midrule
\texttt{cong[A,B,C,D]} & $|AB|/|CD|$ & 1 & Segment equality \\
\texttt{eqratio[A,B,C,D,E,F,G,H]} & $(|AB|/|CD|)/(|EF|/|GH|)$ & 1 & Ratio equality: $AB:CD = EF:GH$ \\
\texttt{eqangle[P\_0,P\_1,P\_2,P\_3,P\_4,P\_5,P\_6,P\_7]} 
& $\angle(P_{0}P_{1},P_{2}P_{3}) - \angle(P_{4}P_{5},P_{6}P_{7})$ 
& 0 & Angle equality (mod 180°) \\
\midrule
\multicolumn{4}{l}{\textit{Parallel and Perpendicular}} \\
\midrule
\texttt{para[A,B,C,D]} & $\angle(AB, CD)$ & 0 & Parallel: $AB \parallel CD$ \\
\texttt{perp[A,B,C,D]} & $\angle(AB, CD)$ & 90 & Perpendicular: $AB \perp CD$ \\
\midrule
\multicolumn{4}{l}{\textit{Circle-Related}} \\
\midrule
\texttt{cyclic[A,B,C,D]} & $\angle(AB,CB) + \angle(AD,CD)$ & 180 & Opposite angles sum to 180° \\
\texttt{on\_circle[X,O,A]} & $|OX|/|OA|$ & 1 & Point $X$ on circle at $O$ \\
\texttt{lc\_tangent[X,A,O]} & $\angle(AX, AO)$ & 90 & Tangent perpendicular to radius \\
\midrule
\multicolumn{4}{l}{\textit{Similarity and Congruence}} \\
\midrule
\texttt{simtrir[A,B,C,D,E,F]} & $\angle(AB,BC) - \angle(DE,EF)$ & 0 & Similar triangles \\
\midrule
\multicolumn{4}{l}{\textit{Collinearity}} \\
\midrule
\texttt{coll[A,B,C]} & $\mathrm{area\_triangle}(A,B,C)$ & 0 & Zero area \\
\texttt{on\_line[X,A,B]} & $\angle(AX, XB)$ & 0 & Point $X$ on line $AB$ \\
\bottomrule
\end{tabular}
\end{adjustbox}

\footnotesize
\vspace{0.6cm}
\caption{Mapping of constant-valued geometric constraints.}
\label{tab:mapping_constants}
\begin{adjustbox}{width=0.78\linewidth}
\begin{tabular}{llcl}
\toprule
\textbf{Predicate} & \textbf{Numeric Form $T$} & \textbf{Expected} & \textbf{Notes} \\
\midrule
\texttt{rconst[A,B,C,X,r]} & $|AB|/|CX|$ & $r$ & Constant ratio $|AB|/|CX| = r$ \\
\texttt{rconst2[X,A,B,r]} & $|AX|/|BX|$ & $r$ & Constant ratio $|AX|/|BX| = r$ \\
\texttt{aconst[A,B,C,X,$\theta$]} & $\angle(AB, CX)$ & $\theta$ & Fixed angle $\theta$ (degrees, mod 180°) \\
\texttt{s\_angle[A,B,X,$\theta$]} & $\angle(AB, BX)$ & $\theta$ & Angle at vertex $B$ equals $\theta$ \\
\texttt{lconst[A,X,l]} & $|AX|$ & $l$ & Fixed length $|AX| = l$ \\
\texttt{midp[M,A,B]} & $|AM|/|MB|$ & 1 & Midpoint: $|AM| = |MB|$ \\
\bottomrule
\end{tabular}
\end{adjustbox}

\footnotesize
\vspace{0.6cm}
\caption{Mapping of special triangle type predicates.}
\label{tab:mapping_triangles}
\begin{adjustbox}{width=0.78\linewidth}
\begin{tabular}{llcl}
\toprule
\textbf{Predicate} & \textbf{Numeric Form $T$} & \textbf{Expected} & \textbf{Notes} \\
\midrule
\texttt{ieq\_triangle[A,B,C]} & $\angle(AB, BC)$ & 60 & Equilateral triangle: all angles 60° \\
\texttt{iso\_triangle[A,B,C]} & $|AB|/|AC|$ & 1 & Isosceles: $|AB| = |AC|$ \\
\texttt{r\_triangle[A,B,C]} & $\angle(AB, AC)$ & 90 & Right triangle: right angle at $A$ \\
\texttt{triangle12[A,B,C]} & $|AB|/|AC|$ & 0.5 & Triangle with $|AB|:|AC| = 1:2$ \\
\texttt{risos[A,B,C]} & $\angle(AB, AC)$ & 90 & Isosceles right triangle at $A$ \\
\texttt{nsquare[X,A,B]} & $\angle(XA, XB)$ & 90 & Isosceles right triangle: $\angle AXB = 90°$ \\
\bottomrule
\end{tabular}
\end{adjustbox}

\footnotesize
\vspace{0.6cm}
\caption{Mapping of quadrilateral and polygon predicates.}
\label{tab:mapping_quads}
\begin{adjustbox}{width=0.78\linewidth}
\begin{tabular}{llcl}
\toprule
\textbf{Predicate} & \textbf{Numeric Form $T$} & \textbf{Expected} & \textbf{Notes} \\
\midrule
\texttt{rectangle[A,B,C,D]} & $\angle(AB, BC)$ & 90 & Rectangle: $\angle ABC = 90°$ \\
\texttt{square[A,B,X,Y]} & $\angle(AB, AX)$ & 90 & Square: right angle + equal sides \\
\texttt{trapezoid[A,B,C,D]} & $\angle(AB, CD)$ & 0 & Trapezoid: $AB \parallel CD$ \\
\texttt{r\_trapezoid[A,B,C,D]} & $\angle(AB, AD)$ & 90 & Right trapezoid: $\angle BAD = 90°$ \\
\texttt{eq\_quadrangle[A,B,C,D]} & $|AD|/|BC|$ & 1 & Quadrilateral with $|AD| = |BC|$ \\
\texttt{eqdia\_quadrangle[A,B,C,D]} & $|AC|/|BD|$ & 1 & Equal diagonals: $|AC| = |BD|$ \\
\texttt{psquare[X,A,B]} & $\angle(AB, AX)$ & 90 & Rotate $B$ by 90° around $A$ to $X$ \\
\bottomrule
\end{tabular}
\end{adjustbox}
\end{table*}

\begin{table*}[t]
\centering
\footnotesize
\caption{Mapping of auxiliary point construction predicates.}
\label{tab:mapping_constructions}

\begin{adjustbox}{width=0.78\linewidth}
\begin{tabular}{llcl}
\toprule
\textbf{Predicate} & \textbf{Numeric Form $T$} & \textbf{Expected} & \textbf{Notes} \\
\midrule
\texttt{on\_pline[X,A,B,C]} & $\angle(AX, BC)$ & 0 & $X$ on parallel to $BC$ through $A$ \\
\texttt{on\_tline[X,A,B,C]} & $\angle(AX, BC)$ & 90 & $X$ on perpendicular to $BC$ through $A$ \\
\texttt{on\_bline[X,A,B]} & $|XA|/|XB|$ & 1 & $X$ on perpendicular bisector of $AB$ \\
\texttt{on\_dia[X,A,B]} & $\angle(AX, BX)$ & 90 & $X$ on circle with diameter $AB$ \\
\texttt{on\_aline[X,A,B,C,D,E]} & $\angle(BA,AX) - \angle(ED,DC)$ & 0 & Angle transfer construction \\
\texttt{reflect[X,A,B,C]} & $\angle(BA,BC) - \angle(CB,CX)$ & 0 & Reflection of $A$ across line $BC$ \\
\texttt{eqangle2[X,A,B,C]} & $\angle(AX,XB) - \angle(CX,XA)$ & 0 & Opposite angles $\angle BXA = \angle CXA$ \\
\texttt{eqangle3[X,A,B,D,E,F]} & $\angle(AX,XB) - \angle(DE,EF)$ & 0 & Angle equality $\angle AXB = \angle DEF$ \\
\texttt{eqratio6[X,A,C,E,F,G,H]} & $(|AX|/|CX|)/(|EF|/|GH|)$ & 1 & Ratio constraint $|AX|:|CX| = |EF|:|GH|$ \\
\bottomrule
\end{tabular}
\end{adjustbox}
\end{table*}

\subsection{Constructed Auxiliary Points}
Numerous predicates encode auxiliary point constructions (e.g., \texttt{on\_pline}, \texttt{on\_tline}) that introduce intermediate geometric entities. Table~\ref{tab:mapping_constructions} defines the corresponding verification conditions.

\section{Training Configurations and Hyperparameters}
\label{appendix:training_configs}

We implement the RLVR training stage using the MM-Eureka~\cite{meng2025mmeurekaexploringfrontiersmultimodal} framework, built upon OpenRLHF~\cite{hu2025openrlhfeasytousescalablehighperformance}. All models, including Qwen2.5-VL-3B and 7B, share an identical hyperparameter configurations. 

Training is conducted on a cluster of 8 NVIDIA H100 GPUs. We utilize DeepSpeed\footnote{\url{https://github.com/microsoft/DeepSpeed}} ZeRO-3 to manage memory efficiency. Following the SGVR framework, we employ Group Relative Policy Optimization (GRPO) to optimize the policy by maximizing the Skeleton Rate (SR). The visual encoder remains frozen during the reinforcement learning process. Key hyperparameters are detailed in Table~\ref{tab:hyperparams}.

\vspace{1cm}
\begin{table}[ht]
\centering
\small
\caption{Hyperparameters for training.}
\begin{tabular}{lc}
\toprule
\textbf{Hyperparameter} & \textbf{Value} \\
\midrule
Base Models & Qwen2.5-VL-3B / 7B Instruct\\
Optimizer & AdamW \\
Actor Learning Rate & $1 \times 10^{-6}$ \\
Training Batch Size & 16 \\
Micro Batch Size & 2 \\
Samples per Prompt ($G$) & 8 \\
KL Coefficient ($\beta$) & 0.01 \\
Max Prompt Length & 4096 \\
Max Generation Length & 4096 \\
Precision & BF16 \\
\bottomrule
\end{tabular}
\label{tab:hyperparams}
\end{table}

\section{Evaluation Details}
\label{app:eval_prompt}

\subsection{Answer Verification Protocol}

To evaluate numerical answer equivalence across diverse mathematical representations (fractions, decimals, radical expressions), we employ GPT-5-nano~\citep{openai2025gpt5systemcard} as an automated equivalence checker. The complete prompt is below:

\begin{tcolorbox}[
    breakable,
    colback=gray!10,
    colframe=gray!50,
    arc=2mm,
    boxrule=0.5pt,
    left=3mm,
    right=3mm,
    top=2mm,
    bottom=2mm
]
\footnotesize
\textbf{Answer Verification Prompt}

Given two solutions from a geometry question, please judge whether the answers are the same or not.

\textbf{Basic Rules:}
\begin{itemize}[leftmargin=*, nosep]
    \item If the two answers are the same, output 1.
    \item If the two answers are different, output 0.
    \item If you are not sure, output NOT SURE.
    \item Do not consider units in the answers, only consider the numerical values.
\end{itemize}

\textbf{Numerical Comparison Rules:}
\begin{enumerate}[leftmargin=*, nosep]
    \item \textbf{Fractions and Decimals:} 1/2 = 0.5, 1/3 = 0.333, 2/3 = 0.667, etc.
    \item \textbf{Floating Point Tolerance:} Numbers are considered the same if they are within 0.02 of each other. Examples: 1.333 and 1.334 are the same.
    \item \textbf{Square Root Expressions:} Convert square roots to decimal approximations for comparison.
    \begin{itemize}[nosep]
        \item $\sqrt{2} \approx 1.414$, $\sqrt{3} \approx 1.732$, $\sqrt{5} \approx 2.236$
        \item For any expression like $a\sqrt{b}$, calculate $a \times \sqrt{b}$ and convert to decimal.
    \end{itemize}
    \item \textbf{Complex Expressions:} For expressions like $(a\sqrt{b})/c$ or fractions involving square roots, convert the entire expression to decimal form.
\end{enumerate}

\textbf{Examples:}
\begin{itemize}[leftmargin=*, nosep]
    \item ``$\sqrt{2}$'' and ``1.414'' $\rightarrow$ Same (\texttt{<evaluation>1</evaluation>})
    \item ``$2\sqrt{2}$'' and ``2.828'' $\rightarrow$ Same (\texttt{<evaluation>1</evaluation>})
    \item ``198/7'' and ``28.29'' $\rightarrow$ Same as ``$20\sqrt{2} \approx 28.28$'' (\texttt{<evaluation>1</evaluation>})
\end{itemize}

\textbf{Task:}

Solution 1: \{solution1\}

Solution 2: \{solution2\}

Please analyze step by step and give your final conclusion in \texttt{<evaluation>result</evaluation>} format.
\end{tcolorbox}

\paragraph{Output parsing.}
The evaluator returns structured judgments in XML format (\texttt{<evaluation>...</evaluation>}). We extract these using regular expressions and encode the results as binary labels (1 for equivalence, 0 for non-equivalence), with ``NOT SURE'' outcomes recorded separately for ambiguous cases that require manual inspection.

\section{Case Studies: Reasoning Failures Leading to Incorrect Answers}
\label{app:case_studies}

We present detailed qualitative comparisons between baseline models and SGVR-trained models across three representative examples spanning probability, logic, and combinatorial reasoning. These cases illustrate systematic differences in constraint adherence and mathematical rigor, where reasoning errors lead to incorrect final answers.

\noindent\textit{Notation:} Baseline responses appear in gray boxes, SGVR-trained responses in blue boxes. Red highlights denote reasoning errors; blue highlights mark corresponding reasoning steps for comparison.

\subsection{Case 1: AMC Dataset -- Ant Amelia Probability Problem}

\paragraph{Problem Statement.}
Ant Amelia starts on the number line at $0$ and crawls in the following manner. For $n=1,2,3$, Amelia chooses a time duration $t_n$ and an increment $x_n$ independently and uniformly at random from the interval $(0,1)$. During the $n$th step of the process, Amelia moves $x_n$ units in the positive direction, using up $t_n$ minutes. If the total elapsed time has exceeded $1$ minute during the $n$th step, she stops at the end of that step; otherwise, she continues with the next step, taking at most $3$ steps in all. What is the denominator plus the numerator of the probability that Amelia's position when she stops will be greater than $1$?

\noindent\textbf{Ground Truth Answer:} 5 (probability is $\frac{2}{3}$)

\vspace{0.5em}
\begin{tcolorbox}[colback=gray!5, colframe=gray!50, arc=2mm, boxrule=0.5pt, left=3mm, right=3mm, top=2mm, bottom=2mm, before skip=0pt, after skip=0pt]
\small
\textbf{Baseline Response:}\\
To solve this problem, we need to determine the probability that Ant Amelia's final position exceeds 1 after at most three steps. We break down the problem into cases based on the number of steps taken.

\textit{[Case 1 and Case 2 analysis omitted: Baseline correctly finds probabilities 0 and $\frac{1}{2}$ respectively.]}

\textbf{Case 3: Three steps}\\
Amelia takes three steps, but only if the first two steps do not cause her to stop early. Her position is $P_3 = x_1 + x_2 + x_3$ if both $t_2 \leq t_1$ and $t_3 \leq t_2$. The condition for her position to exceed 1 is:
\[ x_1 + x_2 + x_3 > 1 \]
\textcolor{red}{This is more complex to analyze directly, so we use symmetry and the fact that the distribution of the sum of uniform random variables is well-known. The probability that the sum of three independent uniform random variables exceeds 1 is known to be $\frac{5}{8}$.}

Combining these probabilities, the overall probability that Amelia's position exceeds 1 is:
\textcolor{red}{\[ 0 + \frac{1}{2} + \frac{5}{8} = \frac{4}{8} + \frac{4}{8} + \frac{5}{8} = \frac{13}{8} \]}
However, since we are asked for the denominator plus the numerator of the simplified fraction representing the probability, we have:
\textcolor{red}{\[ 13 + 8 = 21 \]}

Thus, the answer is:
\[ \boxed{21} \]
\end{tcolorbox}

\vspace{2mm}
\paragraph{Baseline errors.} (1) Directly cites ``known result'' $\frac{5}{8}$ without derivation; (2) Ignores the stopping condition $t_1 + t_2 + t_3 \leq 1$; (3) Incorrect final answer 21.
\vspace{2mm}

\begin{tcolorbox}[colback=blue!5, colframe=blue!50, arc=2mm, boxrule=0.5pt, left=3mm, right=3mm, top=2mm, bottom=2mm, before skip=0pt, after skip=0pt, breakable]
\small
\textbf{SGVR-trained Response:}\\
Let $D_2=x_1+x_2$ and $D_3=x_1+x_2+x_3$. Since $t_1\in(0,1)$ almost surely, Amelia cannot exceed 1 minute during step 1, so $P(\text{stop after step 1})=0$. Moreover,
\textcolor{blue}{
\begin{align*}
\text{stop after step 2} \ &\Longleftrightarrow\ t_1+t_2>1,\\
\text{stop after step 3} \ &\Longleftrightarrow\ t_1+t_2\le 1.
\end{align*}
}
Because $\{t_i\}$ are independent of $\{x_i\}$, \textcolor{blue}{we can factor ``time'' and ``distance'' events.}

\textbf{Case 2: stopping after step 2.}\\
\begin{align*}
P(\text{stop at 2 and }D_2>1)
&=P(t_1+t_2>1)\,P(D_2>1)\\
&=\frac12\cdot\frac12=\frac14.
\end{align*}

\textbf{Case 3: stopping after step 3.}\\
\begin{align*}
P(\text{stop at 3 and }D_3>1)
&=P(t_1+t_2\le 1)\,P(D_3>1)\\
&=\frac12\cdot\frac56=\frac{5}{12}.
\end{align*}
where
\begin{align*}
P(D_3\le 1)
&=\int_0^1 \int_0^{1-x_1} \Bigl(\int_0^{1-x_1-x_2} 1 \, dx_3\Bigr)\,dx_2\,dx_1\\
&=\int_0^1 \int_0^{1-x_1} (1-x_1-x_2)\,dx_2\,dx_1\\
&=\frac16.
\end{align*}

Therefore,
\begin{align*}
P(D>1) &= \frac14+\frac{5}{12}=\frac{2}{3},\\
2+3 &= 5
\end{align*}
\[ \boxed{21} \]
\end{tcolorbox}

\vspace{2mm}
\paragraph{Key distinction.} The SGVR-trained model makes the step-count constraint explicit (step 2 vs.\ step 3 determined by $t_1+t_2$) and uses independence to factor probabilities; in contrast, the baseline ignores the stopping rule and (implicitly) treats the distance sum as unconstrained, leading to an invalid probability $>1$.

\subsection{Case 2: Livebench Reasoning -- Logic Puzzle}

\paragraph{Problem Statement.}
There are 3 people standing in a line numbered 1 to 3. Each person has attributes: Beverage, Food, Movie-Genre, Nationality. Given constraints:
\begin{itemize}[nosep]
    \item Juice drinker is right of soy-milk drinker
    \item Thriller watcher is in even position
    \item Family watcher drinks juice
    \item Apricot eater is right of soy-milk drinker
    \item Pakistani is not left of apricot eater
    \item Grape eater is left of soy-milk drinker
    \item Grape eater is immediately left of British person
\end{itemize}
\textbf{Questions:} (1) Position of Spanish person? (2) Nationality of grape eater? (3) Beverage at position 2? (4) Beverage of family watcher?

\noindent\textbf{Ground Truth Answer:} 1, Spanish, Soy-milk, Juice
\vspace{0.5em}

\begin{tcolorbox}[breakable, colback=gray!5, colframe=gray!50, arc=2mm, boxrule=0.5pt, left=3mm, right=3mm, top=2mm, bottom=2mm, before skip=0pt, after skip=0pt]
\small
\textbf{Baseline Response:}\\
Let's solve these step-by-step using the given constraints.

\textit{[Initial constraint analysis omitted: Baseline correctly identifies that grape-eater and British person have an immediate left-right relationship, and determines that British person is in position 3, grape-eater in position 2.]}

Now, considering the other constraints:
- The person eating apricot must be to the right of the soy-milk drinker.
- The pakistani person cannot be to the left of the apricot eater.

\textcolor{red}{Given the British person is at position 3, the apricot eater must be in position 4 (since they're to the right of the soy-milk drinker).}

So far, we have:
- Position 1: unknown
- Position 2: grape-eater
- Position 3: British
\textcolor{red}{- Position 4: apricot-eater}

\textit{[Remaining attribute assignment steps omitted: Baseline assigns attributes to all positions, including incorrectly assigning Pakistani nationality to position 2 (grape-eater).]}

\textbf{Final Assignment:}\\
- Position 1: Kale-eater (spanish), Juice-drinker, Family-watcher\\
- Position 2: Grapes-eater (pakistani), Soy-milk-drinker, Thriller-watcher\\
- Position 3: British, Lemonade-drinker, Action-watcher\\
- Position 4: Apricot-eater (pakistani), Soy-milk-drinker, Thriller-watcher

\textbf{Answers:}\\
1. At what position is the person who is Spanish? Answer: 1\\
2. What is the nationality of the person who eats grapes? \textcolor{red}{Answer: Pakistani}\\
3. What beverage does the person in position 2 drink? Answer: Soy-milk\\
4. What is the beverage of the person who watches family? Answer: Juice

\textcolor{red}{<solution>1, Pakistani, Soy-milk, Juice</solution>}
\end{tcolorbox}

\vspace{2mm}
\paragraph{Baseline errors.} (1) Assumes position 4 exists when only 3 people are present; (2) Incorrectly assigns Pakistani nationality to grape eater; (3) Final answer incorrectly states grape eater is Pakistani instead of Spanish.

\vspace{0.5em}
\begin{tcolorbox}[colback=blue!5, colframe=blue!50, arc=2mm, boxrule=0.5pt, left=3mm, right=3mm, top=2mm, bottom=2mm, before skip=0pt, after skip=0pt]
\small
\textbf{SGVR-trained Response:}\\
We have only three positions $\{1,2,3\}$. From ``grape eater is immediately left of the British person'', the only possibilities are $(\text{grape},\text{British})=(1,2)$ or $(2,3)$.

\textcolor{blue}{If $(2,3)$, then grape is in position 2, so soy-milk must be in position 3 (grape is left of soy-milk). But then there is no position to the right of soy-milk for juice, contradicting ``juice is right of soy-milk''. Hence $(2,3)$ is impossible.}

Therefore \textcolor{blue}{grape-eater is in position 1 and the British person is in position 2}. Since grape is left of soy-milk, soy-milk cannot be in position 1, and if soy-milk were in position 3 then again no one could be to its right. Thus \textcolor{blue}{soy-milk is in position 2 and juice is in position 3}; by ``family watcher drinks juice'', position 3 watches family.

Finally, apricot is right of soy-milk, so apricot is in position 3; Pakistani is not left of apricot, so Pakistani is also in position 3. The remaining nationality at position 1 is \textcolor{blue}{Spanish}. (Thriller is the even position $\Rightarrow$ position 2.)



\textbf{Answers:}\\
1. At what position is the person who is spanish? \textbf{Answer: 1}\\
2. What is the nationality of the person who eats grapes? \textbf{Answer: Spanish}\\
3. What beverage does the person in position 2 drink? \textbf{Answer: Soy-milk}\\
4. What is the beverage of the person who watches family? \textbf{Answer: Juice}

<solution>1, Spanish, Soy-milk, Juice</solution>
\end{tcolorbox}

\vspace{2mm}
\paragraph{Key distinction.} Both approaches notice the adjacency between grape eater and the British person. However, the SGVR-trained model \textcolor{blue}{uses the 3-position constraint plus ``juice is right of soy-milk'' to rule out the $(2,3)$ adjacency case}, while the baseline invents a non-existent position 4 and derails the remaining assignments.

\subsection{Case 3: Livebench Reasoning -- Heptagon Cutting}

\paragraph{Problem Statement.}
Suppose I have a regular heptagon, and I can make four straight cuts. Each cut cannot pass through any vertices. Also, exactly three of the cuts must intersect at a single point within the heptagon. What is the maximum number of resulting pieces?

\noindent\textbf{Ground Truth Answer:} 10
\vspace{0.5em}
\begin{tcolorbox}[breakable, colback=gray!5, colframe=gray!50, arc=2mm, boxrule=0.5pt, left=3mm, right=3mm, top=2mm, bottom=2mm, before skip=0pt, after skip=0pt]
\small
\textbf{Baseline Response:}\\
To solve this problem, we need to understand how cutting a heptagon with specific conditions affects its division into pieces.

1. \textbf{Initial Setup}: We start with one piece: the entire heptagon.

2. \textbf{First Cut}: The first cut divides the heptagon into two pieces.

3. \textbf{Second Cut}: The second cut intersects the first cut but does not pass through any vertex. This will divide each of the two pieces from the previous step into two more pieces, resulting in $2 + 2 = 4$ pieces.

4. \textbf{Third Cut}: The third cut intersects both of the previous cuts but does not pass through any vertex. This will further divide each of the four pieces into two more pieces, resulting in $4 + 4 = 8$ pieces.

5. \textbf{Fourth Cut}: \textcolor{red}{The fourth cut intersects all three previous cuts but does not pass through any vertex. This will further divide each of the eight pieces into two more pieces, resulting in $8 + 8 = 16$ pieces.}

Thus, after making four such cuts, the maximum number of resulting pieces is $\boxed{16}$.
\end{tcolorbox}

\vspace{2mm}
\paragraph{Baseline errors.} (1) Ignores the constraint that exactly three cuts must intersect at a single point; (2) Incorrectly assumes each cut doubles the region count without accounting for the special intersection constraint; (3) Incorrect final answer 16.

\vspace{0.5em}
\begin{tcolorbox}[breakable, colback=blue!5, colframe=blue!50, arc=2mm, boxrule=0.5pt, left=3mm, right=3mm, top=2mm, bottom=2mm, before skip=0pt, after skip=0pt]
\small
\textbf{SGVR-trained Response:}\\
To solve this problem, we need to understand how cutting a shape with straight lines affects the number of regions it creates. The key insight here is that each new line added can potentially intersect all previous lines, creating additional regions.

Let's break it down step by step:

1. \textbf{Initial Shape}: We start with a heptagon, which has 7 sides.

2. \textbf{First Cut}: When we make the first cut, it will divide the heptagon into two regions. So now we have 2 regions.

3. \textbf{Second Cut}: \textcolor{blue}{The second cut can intersect the first cut and add one more region for each intersection plus one more region if it doesn't intersect the first cut. Since the second cut intersects the first cut once, it adds 2 more regions. So now we have $2 + 2 = 4$ regions.}

4. \textbf{Third Cut}: \textcolor{blue}{The third cut can intersect the first two cuts twice, adding 3 more regions. So now we have $4 + 3 = 7$ regions.}

5. \textbf{Fourth Cut}: The fourth cut can intersect the first three cuts three times, adding 4 more regions. However, since exactly three of the cuts must intersect at a single point, this cut will only add 3 more regions because the intersection point does not create an extra region due to the condition. So now we have $7 + 3 = 10$ regions.

Thus, the maximum number of resulting pieces after making four straight cuts, with exactly three of them intersecting at a single point, is $\boxed{10}$.
\end{tcolorbox}

\vspace{2mm}
\paragraph{Key distinction.} Both approaches correctly analyze cuts 1-3 (yielding 2, 4, and 7 regions). However, for the fourth cut, SGVR-trained model explicitly accounts for the constraint that three cuts intersect at a single point, recognizing this reduces additional regions from 4 to 3, while baseline applies a generic doubling heuristic.

\subsection{Summary: Constraint Adherence and Reasoning Rigor}

A consistent pattern emerges across all three cases: baseline models exhibit plausible initial reasoning but fail to maintain constraint awareness throughout multi-step derivations, while SGVR-trained models maintain systematic constraint verification at every reasoning step.

Specifically, in \textbf{Case 1 (Ant Amelia)}, baseline cites an unsubstantiated result and fails to condition on the stopping rule; SGVR-trained model makes the step-count constraint explicit (step 2 vs.\ step 3 determined by $t_1+t_2$) and computes the distance terms rigorously. In \textbf{Case 2 (Logic Puzzle)}, baseline violates the problem specification by introducing a non-existent position 4; SGVR-trained model systematically verifies constraint compatibility at each inference step. In \textbf{Case 3 (Heptagon Cutting)}, baseline applies a generic counting heuristic without adapting to the special intersection constraint; SGVR-trained model explicitly reasons about how the constraint modifies region generation.

These examples illustrate that \textit{subgoal-level correctness}---maintaining rigorous constraint adherence and mathematical validity throughout the derivation---is essential for reliable problem-solving. SGVR's step-wise verification mechanism ensures logical soundness and constraint compliance at each reasoning step, rather than merely encouraging superficially plausible intermediate steps.

\section{Case Studies: Correct Answers via Incorrect Reasoning}
\label{app:incorrect_reasoning}

We present case studies where baseline models arrive at correct final answers through fundamentally flawed reasoning processes. These examples demonstrate the critical distinction between \textit{outcome correctness} and \textit{subgoal correctness}, illustrating why answer verification alone is insufficient for evaluating mathematical reasoning capabilities.

\subsection{Case 1: AMC Dataset -- 4×4 Matrix Problem}

\paragraph{Problem Statement.}
How many $4 \times 4$ arrays whose entries are $0$s and $1$s are there such that the row sums (the sum of the entries in each row) are $1, 2, 3,$ and $4,$ in some order, and the column sums (the sum of the entries in each column) are also $1, 2, 3,$ and $4,$ in some order? Output the remainder when the answer is divided by 100.

For example, the following array satisfies the condition.
\[
\left[\begin{array}{cccc}
1 & 1 & 1 & 0 \\
0 & 1 & 1 & 0 \\
1 & 1 & 1 & 1 \\
0 & 1 & 0 & 0 \\
\end{array}\right]
\]

\noindent\textbf{Ground Truth Answer:} 76

\vspace{0.5em}
\begin{tcolorbox}[breakable, colback=gray!5, colframe=gray!50, arc=2mm, boxrule=0.5pt, left=3mm, right=3mm, top=2mm, bottom=2mm, before skip=0pt, after skip=0pt]
\small
\textbf{Baseline Response:}\\
To solve this problem, we need to determine how many $4 \times 4$ matrices with entries either 0 or 1 satisfy both the row sums and the column sums being exactly 1, 2, 3, and 4 in some order.

Firstly, note that if the row sums are 1, 2, 3, and 4, then the column sums must be the same because the total number of 1s in the matrix is fixed. This means that the column sums must also be 1, 2, 3, and 4 in some order. The only way this can happen is if each row has exactly one more 1 than the previous row up to the fourth row, which implies that the rows must be arranged in increasing order of their sums from left to right. Therefore, the possible row arrangements are permutations of the sequence $(1, 2, 3, 4)$, and similarly for the columns.

The number of ways to arrange the rows is $4!$ (since any permutation of the four distinct numbers will work). Similarly, the number of ways to arrange the columns is also $4!$. \textcolor{red}{Since these two arrangements are independent, the total number of such matrices is:}
\textcolor{red}{\[
(4!) \times (4!) = 24 \times 24 = 576.
\]}

Finally, we need to find the remainder when 576 is divided by 100:
\[
576 \div 100 = 5 \text{ remainder } 76.
\]
Thus, the remainder when 576 is divided by 100 is:
\[
\boxed{76}.
\]
\end{tcolorbox}

\vspace{2mm}
\paragraph{Reasoning error.} Baseline incorrectly assumes row and column sum arrangements are independent, treating this as a simple permutation problem rather than recognizing the constrained bipartite matching structure.

\paragraph{Analysis.}
The baseline treats row and column sum arrangements as independent ($4! \times 4! = 576$), when matrix entries must simultaneously satisfy both constraints---a combinatorial structure requiring careful enumeration. The correct count is indeed 576, but arrives at this value through fundamentally flawed independence reasoning. The coincidental correctness ($576 \bmod 100 = 76$) demonstrates that outcome-only evaluation fails to detect invalid reasoning paths, underscoring the necessity of subgoal-level verification.

\subsection{Case 2: Olympiad Dataset -- Spherical Geometry Problem}

\paragraph{Problem Statement.}
The Little Prince lives on a spherical planet which has a radius of $24 \mathrm{~km}$ and centre $O$. He hovers in a helicopter $(H)$ at a height of $2 \mathrm{~km}$ above the surface of the planet. From his position in the helicopter, what is the distance, in kilometres, to the furthest point on the surface of the planet that he can see?

\begin{figure}[h]
\centering
\includegraphics[width=0.15\textwidth]{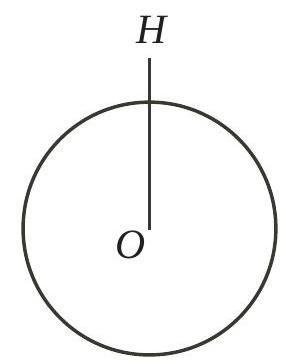}
\caption{Geometric diagram for the spherical geometry problem.}
\label{fig:spherical_geometry}
\end{figure}

\noindent\textbf{Ground Truth Answer:} 10

\vspace{0.5em}
\begin{tcolorbox}[breakable, colback=gray!5, colframe=gray!50, arc=2mm, boxrule=0.5pt, left=3mm, right=3mm, top=2mm, bottom=2mm, before skip=0pt, after skip=0pt]
\small
\textbf{Baseline Response:}\\
To determine the distance to the furthest point on the surface of the planet that the Little Prince can see from his position in the helicopter, we need to analyze the geometry of the situation. The Little Prince is at a height of 2 km above the surface of the planet, and the radius of the planet is 24 km. Therefore, the distance from the center of the planet to the Little Prince is $24 + 2 = 26$ km.

We need to find the distance from the Little Prince's position to the furthest point on the surface of the planet. \textcolor{red}{This distance is the hypotenuse of a right triangle where one leg is the radius of the planet (24 km) and the other leg is the distance from the center of the planet to the Little Prince (26 km).}

Using the Pythagorean theorem, we can find the distance to the furthest point on the surface of the planet. Let $d$ be the distance from the Little Prince to the furthest point on the surface of the planet. Then we have:
\[
d = \sqrt{(26)^2 - (24)^2}
\]
Calculating the squares, we get:
\[
d = \sqrt{676 - 576} = \sqrt{100} = 10 \text{ km}
\]
Therefore, the distance to the furthest point on the surface of the planet that the Little Prince can see is $\boxed{10}$ km.
\end{tcolorbox}

\vspace{2mm}
\paragraph{Reasoning error.} Baseline incorrectly describes the geometry, claiming the sought distance is a triangle hypotenuse with legs (24 km, 26 km)---a geometrically invalid construction for the tangent-line visibility problem.

\paragraph{Analysis.}
The baseline mischaracterizes the geometric configuration: the correct approach identifies the tangent point $T$ on the sphere where the line of sight from $H$ (helicopter) touches the surface. By the tangent-secant relationship, $HT^2 = HO^2 - OT^2 = 26^2 - 24^2 = 100$, yielding $HT = 10$ km. Although the numerical calculation fortuitously produces the correct answer, the underlying geometric reasoning is invalid. This case exemplifies how answer-only evaluation can fail to detect conceptual errors, reinforcing the value of step-wise subgoal verification.

\end{document}